\documentclass{article}
\usepackage{microtype}
\usepackage{graphicx}
\usepackage{subcaption}
\usepackage{booktabs} 
\usepackage{multirow}
\usepackage{makecell}
\usepackage{adjustbox}
\usepackage{etoc}        
 \usepackage{fancyvrb}
\usepackage{threeparttable}
\usepackage{caption}
\usepackage[skins,breakable,listings]{tcolorbox}
\tcbuselibrary{skins,breakable,listings}

\usepackage{hyperref}
\usepackage{enumitem}

\usepackage[preprint]{icml2026}

\usepackage{amsmath}
\usepackage{amssymb}
\usepackage{mathtools}
\usepackage{amsthm}

\usepackage[capitalize,noabbrev]{cleveref}

\theoremstyle{plain}
\newtheorem{theorem}{Theorem}[section]

\theoremstyle{definition}

\theoremstyle{remark}

\usepackage[textsize=tiny]{todonotes}

\icmltitlerunning{MemRL: Self-Evolving Agents via Runtime Reinforcement Learning on Episodic Memory}

\begin{document}
\twocolumn[
  
  \icmltitle{MemRL: Self-Evolving Agents via Runtime Reinforcement Learning on
  \\Episodic Memory}



\icmlsetsymbol{equal}{*}

\begin{icmlauthorlist}
  \icmlauthor{Shengtao Zhang}{sjtu,equal}
  \icmlauthor{Jiaqian Wang}{xd,equal}
  \icmlauthor{Ruiwen Zhou}{nus}
  \icmlauthor{Junwei Liao}{sjtu,sii}
  \icmlauthor{Yuchen Feng}{mem}
  \icmlauthor{Zhuo Li}{sjtu}
  \icmlauthor{Yujie Zheng}{sjtu}
  \icmlauthor{Weinan Zhang}{sjtu,sii}
  \icmlauthor{Ying Wen}{sjtu,sii}
  \icmlauthor{Zhiyu Li}{mem}
  \icmlauthor{Feiyu Xiong}{mem}
  \icmlauthor{Yutao Qi}{xd}
  \icmlauthor{Bo Tang}{ustc,mem}
  \icmlauthor{Muning Wen}{sjtu}
\end{icmlauthorlist}

\icmlaffiliation{sjtu}{Shanghai Jiao Tong University, Shanghai, China}
\icmlaffiliation{xd}{Xidian University, Xi'an, China}
\icmlaffiliation{nus}{National University of Singapore, Singapore}
\icmlaffiliation{sii}{Shanghai Innovation Institute, Shanghai, China}
\icmlaffiliation{mem}{MemTensor (Shanghai) Technology Co., Ltd., Shanghai, China}
\icmlaffiliation{ustc}{University of Science and Technology of China, Hefei, China}

\icmlcorrespondingauthor{Bo Tang}{tangb@memtensor.cn}
\icmlcorrespondingauthor{Muning Wen}{muningwen@sjtu.edu.cn}
  \icmlkeywords{Machine Learning, ICML}


        


  \vskip 0.3in
]



\printAffiliationsAndNotice{\icmlEqualContribution}

\begin{abstract}
The hallmark of human intelligence is the self-evolving ability to master new skills by learning from past experiences. However, current AI agents struggle to emulate this self-evolution: fine-tuning is computationally expensive and prone to catastrophic forgetting, while existing memory-based methods rely on passive semantic matching that often retrieves noise. To address these challenges, we propose \textsc{MemRL}, a non-parametric approach that evolves via reinforcement learning on episodic memory. By decoupling stable reasoning from plastic memory, \textsc{MemRL} employs a Two-Phase Retrieval mechanism to filter noise and identify high-utility strategies through environmental feedback. Extensive experiments on HLE, BigCodeBench, ALFWorld, and Lifelong Agent Bench demonstrate that \textsc{MemRL} significantly outperforms state-of-the-art baselines, confirming that \textsc{MemRL} effectively reconciles the stability-plasticity dilemma, enabling continuous runtime improvement without weight updates. Code is available at \url{https://github.com/MemTensor/MemRL}.
\end{abstract}

\section{Introduction}

Human intelligence balances cognitive stability and episodic plasticity \citep{grossberg2013adaptive,mcclelland1995there,kumaran2016learning} via Constructive Episodic Simulation, enabling adaptation without rewiring neural circuitry \citep{schacter2007constructive,hassabis2007deconstructing,schacter2012future,gick1980analogical}. Despite their reasoning capabilities, current AI agents struggle to emulate this decoupled self-evolution \citep{wei2022chain,yao2022react,schick2023toolformer,wang2023voyager}. Specifically, fine-tuning internalizes experience by modifying weights \citep{ouyang2022training,stiennon2020learning,rafailov2023direct,ethayarajh2024kto} but suffers from computational costs and catastrophic forgetting \citep{kirkpatrick2017overcoming,li2024revisiting,wu2024continuallearninglargelanguage}. Conversely, Retrieval-Augmented Generation (RAG) \citep{lewis2020retrieval} provides a non-parametric alternative but remains passive, retrieving by semantic similarity rather than utility \citep{karpukhin2020dense,gao2023retrieval}; this prevents agents from effectively leveraging runtime feedback to distinguish high-value strategies from noise.

\begin{figure}[h!]
    \centering
    \includegraphics[width=\linewidth]{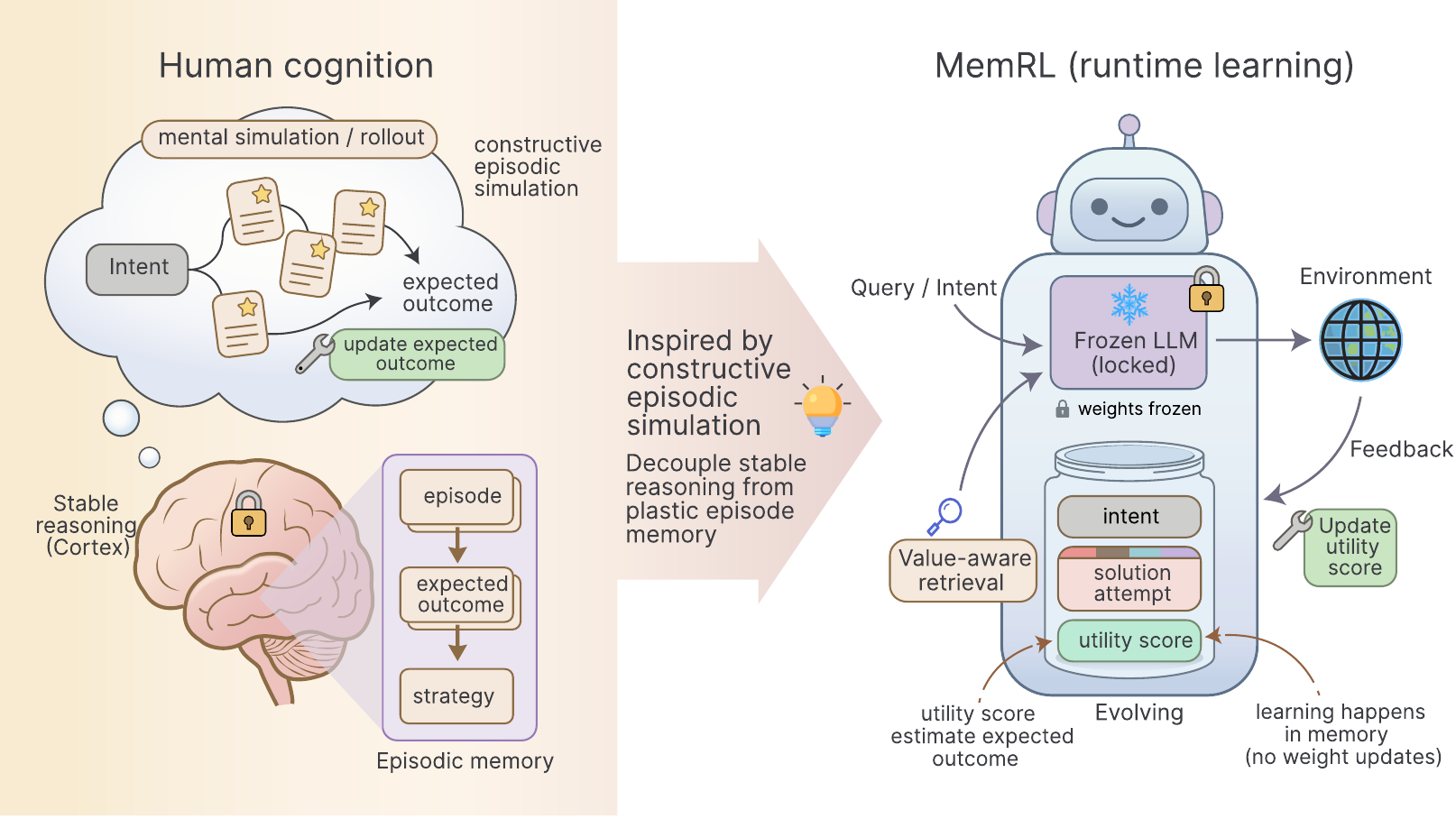} 
    \caption{The conceptual framework of MemRL.}
    \label{fig:memrl_framework}
\end{figure}

This limitation underscores a critical research question: \textbf{How can we enable an agent to continuously improve its performance after deployment, without compromising the stability of its pre-trained backbone?} Our objective is to achieve an agent that evolves with continued usage and rapidly adapts to new tasks after deployment, referred to as Runtime Continuous Learning  \citep{javed2023online,silver2025era_of_experience,parisi2019continual,wu2024continuallearninglargelanguage}, all while keeping the backbone model frozen to prevent catastrophic forgetting \citep{finn2017model,wei2025evo}.
To address this challenge, inspired by the human cognitive mechanism of constructive simulation, we propose \textbf{\textsc{MemRL}}, an approach that facilitates self-evolving agents by explicitly decoupling the model’s stable cognitive reasoning from dynamic episodic memory. Figure~\ref{fig:memrl_framework} illustrates the conceptual framework of our proposed \textsc{MemRL}.
Drawing on tries-and-errors manner in Reinforcement Learning (RL) to estimate expected experience utilities  \citep{sutton2018reinforcement}, we formalize the interaction between the frozen LLM and external memory as a Markov Decision Process (MDP)  \citep{puterman2014markov}. Unlike traditional methods that optimize the backbone model, \textsc{MemRL} optimizes the policy of memory usage without tuning model weights. 

\textsc{MemRL} organizes memory into a structured \textbf{Intent-Experience-Utility} triplet. This structure transforms retrieval from a passive semantic match task into an active decision-making process: \textbf{Two-Phase Retrieval} selects experiences based on their learned Q-values, reflecting expected utility, rather than semantic similarity alone \citep{watkins1992q}; \textbf{Utility-Driven Update} refines these Q-values through environmental feedback, applying Monte Carlo style updates~\citep{metropolis1949monte}.
This closed-loop cycle enables the agent to distinguish high-value memories from similar noise, effectively learning from both success and failure without high computational cost or catastrophic forgetting risks associated with weight updates. As for experiments, we validate \textsc{MemRL} on four diverse benchmarks, including HLE, BigCodeBench, ALFWorld, and Lifelong Agent Bench. Our results demonstrate consistent superiority over baselines, achieving relative improvement in exploration-heavy environments. Our in-depth analysis reveals a strong correlation between learned utility and task success, further confirming \textsc{MemRL}’s effectiveness.

In summary, our contributions are threefold:
\begin{itemize}
    \item We propose a runtime learning framework using Model-Memory decoupling and Intent-Experience-Utility triplet to reconcile the stability-plasticity dilemma, enabling tuning-free agent learning.
\item We introduce \textsc{MemRL}, a non-parametric approach enabling agent self-evolution via Two-Phase Retrieval and Utility-Driven Update.
    \item We conduct extensive evaluations and provide a rigorous analysis for \textsc{MemRL}’s stability, showing how it ensures task integrity and minimizes forgetting.
\end{itemize}


\section{Related Works}
\paragraph{Runtime Learning}
Runtime Learning focuses on the post-deployment improvement of agents through interaction streams rather than offline data, marking a shift toward the ``era of experience'' \citep{silver2025era_of_experience}. Unlike Continual Learning \citep{parisi2019continual,wu2024continuallearninglargelanguage} or Test-Time Adaptation \citep{sun2020test,wang2020tent,liang2025comprehensive}, which typically update parameters to handle forgetting or distribution shifts, our setting constrains the backbone to remain frozen to ensure stability and efficiency. While recent memory-augmented agents \citep{zheng2025lifelong,wei2025evo,zhou2025rulearena} emphasize memory organization, the selection problem---identifying which experiences to reuse under feedback---remains a critical challenge. Drawing from value-aware episodic control \citep{tulving1972episodic,mcclelland1995there}, we frame runtime learning as identifying valuable episodes. By using interaction feedback to assign utility, our approach guides retrieval and reuse without weight modification, thereby ensuring sustained improvement \citep{pritzel2017neural}.

\paragraph{Reinforcement Learning}

Reinforcement learning has been widely adopted for LLMs enhancement. A representative paradigm is to construct reward signals from human feedback and optimize the model policy accordingly to align with human preference  \citep{stiennon2020learning,ouyang2022training}. Other recent approaches leverage rule-based verifiers to improve LLMs' reasoning capabilities  \citep{guo2025deepseek,yu2025dapo}. In parallel, agent-oriented research explores how interaction signals can improve tool use and action decision-making, and investigates mechanisms by which language models execute composite actions in environments  \citep{schick2023toolformer}. Despite the demonstrated effectiveness of reward-driven optimization, these methods generally place learning in the model parameters or additional parametric modules, and thus do not avoid the cost of online updates or the risk of forgetting. In contrast, our method frames memory usage as a learnable decision problem and applies \emph{non-parametric} reinforcement learning on memory to bypass the risk.

\paragraph{Agentic Memory}

\begin{figure*}[h]
    \centering
    \includegraphics[width=\linewidth]{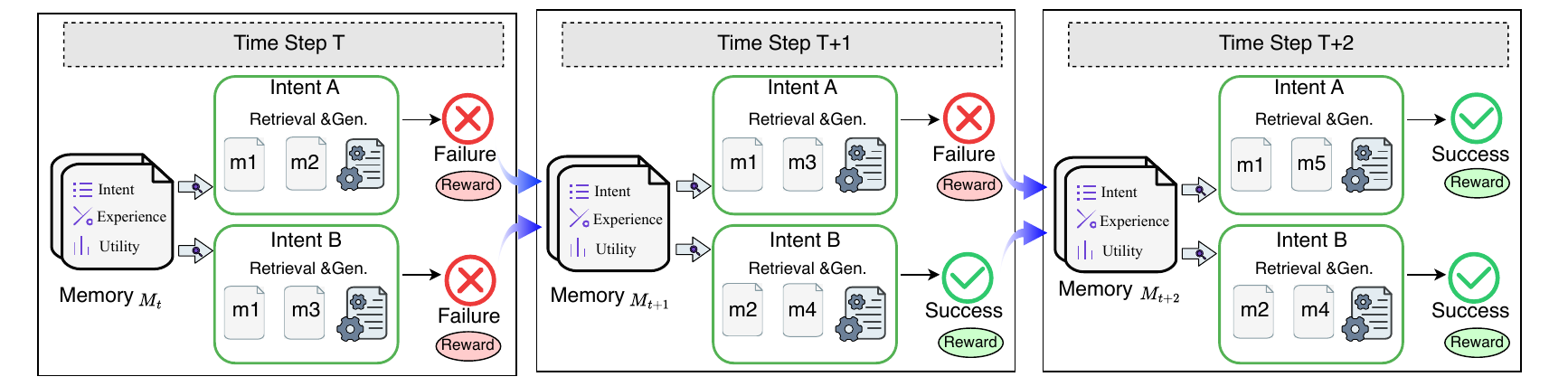}
    \caption{
    An illustrative example of memory-augmented decision making under a Markov Decision Process.
    At time step $t$, the agent starts with an initial memory set $\mathcal{M}_t$.
    At time step $t{+}1$, an intent (\textit{Intent A}) retrieves relevant past experiences, but initially leads to a failed generation.
    In contrast, another intent (\textit{Intent B}) succeeds, and its associated experience is added to memory.
    At time step $t{+}2$, \textit{Intent A} retrieves the newly stored successful experience from \textit{Intent B}, resulting in a successful outcome.
    This example shows how memory retrieval enables knowledge reuse across intents, implicitly supporting self-evolution through shared experiences.
    }
    \label{fig:mmdp}
\end{figure*}

To avoid the costs of fine-tuning, external memory systems have evolved from a static RAG paradigm to dynamic, governable memory structures  \citep{lewis2020retrieval,karpukhin2020dense}. Early agentic memory introduced reflection mechanisms and hierarchical management to handle long context experiences \citep{shinn2023reflexion,packer2024memgptllmsoperatingsystems}. More recent frameworks have systematized the memory lifecycle, focusing on unified storage and structured indexing for complex tasks  \citep{li2025memos,xu2025mem,huang2025licomemory,ye2025task}. Furthermore, adaptive approaches now explore improving retrieval via feedback-driven updates or automated augmentation  \citep{salama2025meminsight,zhang2025learn,li2025retrieval,zhou2025memento}. However, except for training additional learnable modules, most existing methods still rely predominantly on semantic similarity or heuristic rules, lacking a rigorous metric to evaluate the actual utility of a memory in maximizing returns. Inspired by cognitive theories of memory reconsolidation  \citep{schacter2007constructive,gick1980analogical,nader2000fear}, \textsc{MemRL} bridges this gap by formulating retrieval as a value-based decision process, learning robust utility estimates (Q-values) from environmental rewards to distinguish high-value experiences.

\section{Problem Formulation}
\label{sec:problem}

In this section, we formally define the problem of memory-augmented generation and establish the theoretical link between agent policy and memory retrieval. We adopt the formulation of Memory-Based Markov Decision Process (M-MDP)  \citep{zhou2025memento}, and apply our non-parametric reinforcement learning approach to it.
Figure~\ref{fig:mmdp} provides an illustrative example of this memory-augmented decision process, showing how retrieval outcomes and memory evolution unfold over multiple time steps.

\subsection{Memory-Augmented Agent Policy}
\label{sec:mmdp}
To enable the agent to self-evolution, we adopt the M-MDP framework \citep{zhou2025memento}, defined by the tuple $(\mathcal{S}, \mathcal{A}, P, \mathcal{R}, \gamma, \mathcal{M})$. Here, $\mathcal{S}$ and $\mathcal{A}$ represent state and action spaces, $P$ is the transition dynamics, $\mathcal{R}$ is the reward function of state and action, $\gamma \in [0,1)$ denotes the discount factor, and $\mathcal{M} = (\mathcal{S} \times \mathcal{A} \times \mathbb{R})^*$ constitutes the evolving memory bank of past experiences \citep{zhou2025memento}. At each step $t$, the agent receives state $s_t$ and leverages $\mathcal{M}_t$ to generate a response $a_t$ maximizing the expected reward. The joint policy $\pi(a_t | s_t, \mathcal{M}_t)$ is formulated as the marginal probability over all possible retrieved items $m$ \citep{zhou2025memento}:
\begin{equation}
\pi(a_t | s_t, \mathcal{M}_t) = \sum_{m \in \mathcal{M}_t} \mu(m | s_t, \mathcal{M}_t) p_{LLM}(a_t | s_t, m) \label{eq:joint_policy}.
\end{equation}
where $\mu(m | s_t, \mathcal{M}_t)$ is the \textbf{Retrieval Policy} for selecting memory contexts, and $p_{LLM}(a_t | s_t, m)$ is the \textbf{Inference Policy} parameterized by a frozen LLM. This approach transforms retrieval from a passive match into an active decision process, effectively accounting for the functional \emph{utility} of $m$ in generating successful outcomes $a_t$.

In previous RAG or memory-based agentic paradigms, the retrieval policy $\mu$ is usually determined by a fixed vector similarity metric, e.g., cosine similarity of embeddings. While effective for semantic matching, such policies fail to account for the \textit{utility} of a memory, i.e., whether retrieving $m$ actually leads to a successful outcome $a_t$.

\begin{figure*}[h]
    \centering
    \includegraphics[width=\textwidth]{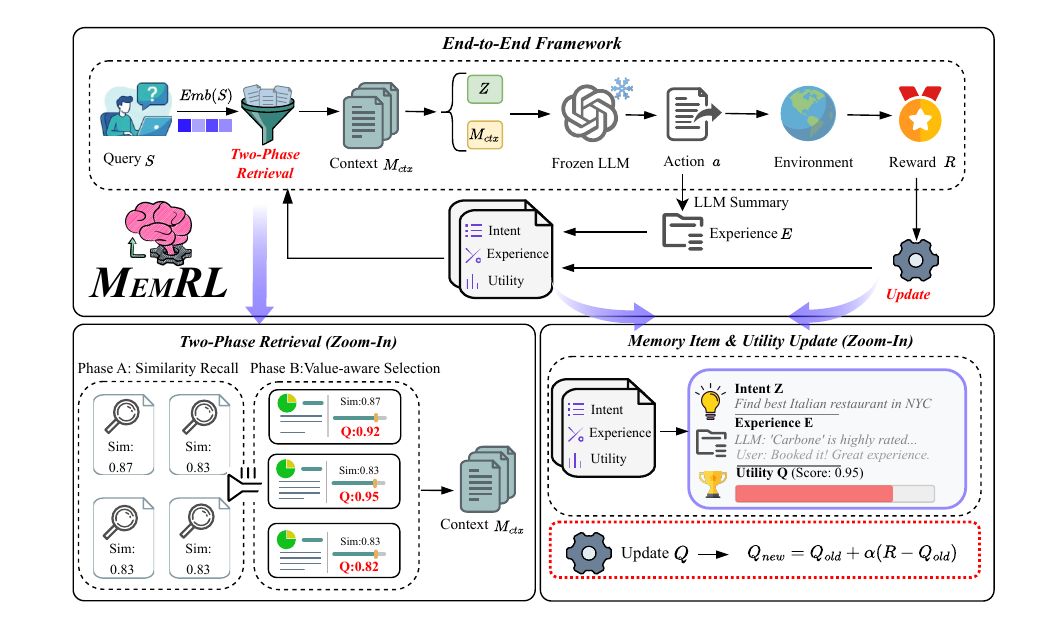} 
    
    \caption{\textbf{Overview of \textsc{MemRL}.} 
    \textbf{The end-to-end learning loop(Top)}: given a query $\mathbf{s}$, the agent retrieves context $\mathbf{m}_{ctx}$ from memory $\mathbf{M}$, generates output $\mathbf{y}$, and updates memory value $Q$ based on reward $R$.
    \textbf{Two-Phase Retrieval(Bottom Left)}: Candidates are recalled via similarity, then re-ranked using learned Q-values.
    \textbf{Utility Update(Bottom Right)}: Values ($Q$) are updated using environmental rewards to distinguish functional utility from semantic similarity.} 
    
    \label{fig:framework}
\end{figure*}

\subsection{Non-Parametric Reinforcement Learning}
To overcome static similarity limitations, we operationalize the M-MDP framework by formulating memory retrieval as a value-based decision-making process \citep{zhou2025memento}. Unlike parametric methods optimizing $\pi_{LLM}$ via weight updates, we optimize the retrieval policy $\mu(m|s, \mathcal{M})$ directly within the memory space by mapping M-MDP components to a structured \textit{Intent-Experience-Utility} triplet:

\textbf{From Semantic Matching to Decision Making.} We instantiate state $s$ as the User Intent, encapsulated by the current query embedding \citep{lewis2020retrieval}. Consequently, the action space $\mathcal{A}_t$ becomes dynamic and discrete, corresponding to selecting a specific $m$ from the memory bank $\mathcal{M}_t$ \citep{zhou2025memento}. In this formulation, retrieval is not a passive matching task but a strategic decision step taken to augment the generator's action $a$~\citep{zhou2025memento}.

\textbf{Defining Utility via Q-Values.} While the agent's executable action $a$ is generated by the policy $\pi$ (as formalized in Sec.~\ref{sec:mmdp}), the quality of this generation is strictly conditioned on the retrieved context. Therefore, we adapt the traditional value function $Q(s, a)$ to the retrieval phase, defining $Q(s, m)$ as the expected utility of the subsequent action $a$ augmented by memory $m$. \textsc{MemRL} learns an optimal retrieval policy $\mu^*$ that maximizes this utility:
\begin{equation}
    \mu^*(m| s, \mathcal{M}) = \arg\max_{m \in \mathcal{M}} Q(s, m). \label{equ:optimal-policy}
\end{equation}
In this view, the Q-value acts as a critic for the retrieval mechanism, distinguishing memories that strategically aid the generator from irrelevant noise that merely shares high semantic similarity.


\textbf{Non-Parametric Learning.} Since the retrieval action space is decoupled from LLM generation, we perform learning without modifying model weights. Upon receiving environmental feedback $r$, we update the Q-value via a Temporal-Difference (TD) error \citep{sutton1988learning}:
\begin{equation}
\label{equ:bellman-q}
    Q(s,m) \leftarrow Q(s,m) + \alpha [r + \gamma \max Q(s',m') - Q(s,m)],
\end{equation}
or Monte Carlo style rule \citep{metropolis1949monte}:
\begin{equation}
\label{eq:q_update}
Q_{\text{new}} \leftarrow Q_{\text{old}} + \alpha \big(r - Q_{\text{old}}\big),
\end{equation}
where $\alpha$ is the learning rate. Equation~\ref{eq:q_update} performs as a naturally simplified version of Equation~\ref{equ:bellman-q} by setting $s'$ as a terminal state to balance complexity and performance, sharing a similar one-step MDP formulation with \citet{guo2025deepseek}. These manners allow utility estimates to converge to true expected returns over time \citep{bellman1966dynamic}. By explicitly updating Q-values within the memory structure, \textsc{MemRL} provides a non-parametric learning manner with a theoretical guarantee, enabling agents to self-evolve through interaction.

\section{\textsc{MemRL}}
\label{sec:method}


Building upon the M-MDP formulation defined in Section~\ref{sec:problem}, we propose \textsc{MemRL}, a framework that enables frozen LLMs to self-evolve via non-parametric reinforcement learning. Instead of modifying the model weights $\theta$, \textsc{MemRL} optimizes the \textbf{retrieval policy} $\mu(m| s, \mathcal{M})$ within an evolving memory space. As illustrated in Figure~\ref{fig:framework}, the framework consists of three core components: (i) a structured \textbf{Intent-Experience-Utility} memory bank, (ii) a \textbf{Two-Phase Retrieval} mechanism that decouples semantic recall from value-aware selection, and (iii) a \textbf{Runtime Utility Update} rule that stabilizes Q-value estimation.



\subsection{The Intent-Experience-Utility Triplet}
\label{subsec:memory_repr}



To support value-based decision-making, we structure the external memory $\mathcal{M}$ not merely as key-value pairs, but as a set of triplets:
\begin{equation}
\mathcal{M}=\{(z_i, e_i, Q_i)\}_{i=1}^{|\mathcal{M}|},
\end{equation}
where $z_i$ represents the \textbf{Intent}, $e_i$ stores the raw \textbf{Experience} (e.g., a successful solution trace or trajectory), and $Q_i$ denotes the learned \textbf{Utility}. $Q_i$ approximates the expected return of applying experience $e_i$ to intents similar to $z_i$, serving as the \textit{critic} in RL.

\subsection{From Semantic Recall to Value-Aware Selection}
\label{subsec:two_phase}
Standard RAG systems assume ``similar implies useful,'' but agentic tasks often involve environment-specific routines that generalize poorly \citep{singh2025agentic,Cuconasu_2024,gan2024similarity,zhou2024trad}. Therefore, \textsc{MemRL} implements a \textbf{Two-Phase Retrieval} strategy. 
\textbf{Phase A: Similarity-Based Recall.} Given query $s$, we isolate a candidate pool $\mathcal{C}(s)$ of semantically consistent experiences by filtering memory bank $\mathcal{M}$ via cosine similarity and a sparsity threshold $\delta$:
\begin{equation}
\label{eq:phase a}
\mathcal{C}(s) = \text{TopK}_{k_1}(\{i | sim(Emb(s), Emb(z_i)) > \delta\}),
\end{equation}
where $Emb$ represents the Embedding Model to transfer the raw text to a vector.
If $\mathcal{C}(s)=\emptyset$, \textsc{MemRL} relies solely on the frozen LLM for exploration. 
\textbf{Phase B: Value-Aware Selection.} To determine the final context $\mathcal{M}_{ctx}(s)$, we select top-$k_2$ items from $\mathcal{C}(s)$ using a composite score balancing exploration (similarity) and exploitation (utility $Q$):
\begin{equation}
\label{eq:score}
\text{score}(s, z_i, e_i) = (1-\lambda) \cdot \hat{sim}(Emb(s), Emb(z_i)) + \lambda \cdot \hat{Q_i}.
\end{equation}
where $\hat{\cdot}$ denotes z-score normalization and $\lambda \in [0, 1]$ modulates the trade-off. This mechanism filters out ``distractor'' memories—those semantically similar but with low historical utility. As detailed in Section~\ref{sec:stable in mem rl exp}, normalization and strict similarity thresholds are essential for noise filtering and maintaining stability during self-evolution.

\subsection{Non-Parametric RL on Memory}
\label{subsec:update}

The core of \textsc{MemRL} is the continuous refinement of Q-values based on environmental feedback, enabling the agent to ``remember'' what works. During runtime, \textsc{MemRL} performs learning entirely in memory space. With the retrieved context $m$, the agent samples an action $a$ according to the policy $\pi$ defined in Eq.~\ref{eq:joint_policy}. Executing $a$ then yields an environmental reward $r$ (e.g., execution success or scalar score). For the memories actually injected into the input context $\mathcal{M}_{\text{ctx}}(s)$, we update their utilities in triplets with a Monte Carlo style rule, i.e., the Eq.~\ref{eq:q_update}, following the runtime learning loop shown in Figure~\ref{fig:framework}. 
This update drives $Q_{\text{new}}$ toward the empirical expected return of using experience $e_i$ under similar intents. Meanwhile, for each sampled trajectory, we use an LLM to summarize the experience~\citep{fang2025memp}, and write it back into the memory bank as a new triplet $(z, e_{\text{new}}, Q_{\text{init}})$, enabling continual expansion of experience while keeping the LLM parameters unchanged.



\subsection{Theoretical Stability Analysis}
\label{sec:stability}

We analyze the stability of \textsc{MemRL} from a reinforcement learning perspective, with full analysis provided in Appendix~\ref{app:theoretical_analysis}. We posit two standard assumptions: a frozen inference policy $p_{\mathrm{LLM}}$ and a stationary task distribution. Under these conditions, the learning target $\beta(s,m) = \mathbb{E}[r_t | s, m]$ is well-defined, where the expectation is taken over the stochastic rewards resulting from the Inference Policy $p_{LLM}$ distribution. We prove that utility estimates updated via Eq.~\ref{eq:q_update} are unbiased and variance-bounded \citep{sutton2018reinforcement}. Specifically, as $t \to \infty$:
\begin{equation}
\label{eq:stability_summary}
\begin{aligned}
\lim_{t \to \infty} \mathbb{E}[Q_t] &= \beta(s,m), \\
\limsup_{t \to \infty} \mathrm{Var}(Q_t) 
&\le \frac{\alpha}{2-\alpha}\mathrm{Var}(r_t \mid s,m).
\end{aligned}
\end{equation}

Furthermore, we address the challenge of the latent retrieval distribution $\Pr(s|m)$ shifting during training by framing \textsc{MemRL} as a \textbf{Generalized Expectation-Maximization (GEM)}~\citep{dempster1977maximum} process. The system performs coordinate ascent on a global objective: the retrieval ranking acts as the \textit{Policy Improvement} (E-step), while the utility update acts as the \textit{Value Update} (M-step). By the \textit{monotonic improvement theorem} \citep{neal1998view}, the system converges to a stationary point where the global memory utility stabilizes:
\begin{equation}
\lim_{t \to \infty} \mathbb{E}[Q_t(m)] 
= \sum_{s \in \mathcal{S}(m)} \mathbb{E}[r | s, m] \Pr(s | m).
\end{equation}
where $\mathcal{S}(m)$ is the effective support set of the memory. This formulation guarantees global stability and prevents catastrophic forgetting. Details can be found in Appendix~\ref{app:gem_analysis}.

\section{Experiments}
\subsection{Experimental Setup}

\label{sec:setup}

\textbf{Baselines \& Benchmarks.} We compare MemRL against RAG-based (RAG, Self-RAG), Agentic Memory (Mem0, MemP), and Test-Time Scaling (Pass@$k$) baselines under a frozen-backbone setting (see Appendix \ref{app:baselines}). Evaluations span four domains: BigCodeBench (coding), ALFWorld (navigation), LifelongAgent Bench (OS/DB), and Humanity's Last Exam(HLE). Details are in Appendix \ref{app:benchmark}. Backbones are selected per benchmark to avoid no-signal or ceiling problems, ensuring valid learning signals, while Appendix~\ref{subsec:model_selection_regimes} provides a unified comparison to demonstrate cross-task consistency under identical capacity. 

\textbf{Metrics.} We employ two metrics: (1) \textbf{Success Rate (SR)}, the ratio of tasks completed in an epoch; (2) \textbf{Cumulative Success Rate (CSR)}, the proportion of tasks solved at least once across epochs.

We evaluate our \textsc{MemRL} and baselines under two distinct settings: \textbf{Runtime Learning}, which assesses the ability to learn and adapt within a training session, and \textbf{Transferring}, which evaluates the generalization capability of the learned memory on unseen tasks. Implementation and reproducibility details, including all prompts used in our experiments, can be found in Appendix~\ref{app:impl} and Appendix~\ref{app:prompts}.
\begin{table*}[t]
\centering
\footnotesize 
\setlength{\tabcolsep}{3pt} 
\setlength{\abovecaptionskip}{3pt}

\caption{\textbf{Runtime Learning results.} We compare \textsc{MemRL} against various baselines over 10 epochs. The results are reported as \textbf{Last Epoch Success Rate / Cumulative Success Rate (CSR)}. The \textbf{Average} column indicates the mean performance across all benchmarks.}
\label{tab:runtime_main_results}

\begin{tabular}{l c c c c c c}
\toprule
& \textbf{BigCodeBench} & \multicolumn{2}{c}{\textbf{Lifelong Agent Bench}} & \textbf{ALFWorld} & \textbf{HLE} & \textbf{Average} \\
\cmidrule(lr){2-2} \cmidrule(lr){3-4} \cmidrule(lr){5-5} \cmidrule(lr){6-6} \cmidrule(lr){7-7}
\textbf{Method} & \begin{tabular}{@{}c@{}}Code Gen\\(Last / CSR)\end{tabular} & 
                  \begin{tabular}{@{}c@{}}OS Task\\(Last / CSR)\end{tabular} & 
                  \begin{tabular}{@{}c@{}}DB Task\\(Last / CSR)\end{tabular} & 
                  \begin{tabular}{@{}c@{}}Exploration\\(Last / CSR)\end{tabular} & 
                  \begin{tabular}{@{}c@{}}Knowledge Frontier\\(Last / CSR)\end{tabular} &
                  \begin{tabular}{@{}c@{}}(Last / CSR)\end{tabular} \\ 
\cmidrule(lr){1-7} 
\textit{Model} & \multicolumn{1}{c}{\texttt{GPT-4o}} & \multicolumn{1}{c}{\texttt{GPT-4o-mini}} & \multicolumn{1}{c}{\texttt{GPT-4o-mini}} & \multicolumn{1}{c}{\texttt{GPT-5-mini}} & \multicolumn{1}{c}{\texttt{Gemini-3-pro}} & - \\
\midrule
No Memory       & 0.485 & 0.674 & 0.860  & 0.777  & 0.357  & 0.631 \\
Pass@10          & -- / 0.577  & -- / 0.756  & -- / 0.928  & -- / 0.928 & -- / 0.524  & -- / 0.743 \\
RAG             & 0.475 / 0.483 & 0.690 / 0.700 & 0.914 / 0.916 & 0.887 / 0.930  & 0.430 / 0.475 & 0.679 / 0.699 \\
Self-RAG        & 0.497 / 0.561 & 0.646 / 0.732 & 0.891 / 0.898 & 0.907 / 0.962 & 0.423 / 0.475  & 0.673 / 0.726 \\
Mem0            & 0.487 / 0.495 & 0.670 / 0.702 & 0.920 / 0.926 & 0.894 / 0.969 & 0.436 / 0.470  & 0.681 / 0.712 \\
MemP            & 0.578 / 0.602 & 0.736 / 0.742 & \textbf{0.960} / 0.966 & 0.885 / 0.919 & 0.522 / 0.570 & 0.736 / 0.760 \\
\midrule
\textit{MemRL (ours)}       & \textbf{0.595 / 0.627} & \textbf{0.788 / 0.804} & \textbf{0.960 / 0.972} & \textbf{0.949 / 0.981} & \textbf{0.570 / 0.606} & \textbf{0.772 / 0.798} \\
\bottomrule
\end{tabular}
\end{table*}

\begin{table*}[t]
\centering
\small 
\setlength{\tabcolsep}{8pt} 
\setlength{\abovecaptionskip}{3pt}

\caption{\textbf{Transfer Learning results} on BigCodeBench, Lifelong Agent Bench and ALFWorld. We compare \textsc{MemRL} against various baselines using the best validation results. The \textbf{Average} column represents the mean Success Rate across all benchmarks.}
\label{tab:transfer_results} 

\begin{tabular}{l c c c c c}
\toprule
& \textbf{BigCodeBench} & \multicolumn{2}{c}{\textbf{Lifelong Agent Bench}} & \textbf{ALFWorld} & \textbf{Average} \\
\cmidrule(lr){2-2} \cmidrule(lr){3-4} \cmidrule(lr){5-5} \cmidrule(lr){6-6}
\textbf{Method} & \begin{tabular}{@{}c@{}}Code Generation\\(Success Rate)\end{tabular} & 
                  \begin{tabular}{@{}c@{}}OS Task\\(Success Rate)\end{tabular} & 
                  \begin{tabular}{@{}c@{}}DB Task\\(Success Rate)\end{tabular} & 
                  \begin{tabular}{@{}c@{}}Exploration\\(Success Rate)\end{tabular} &
                  \begin{tabular}{@{}c@{}}(Success Rate)\end{tabular} \\ 
\cmidrule(lr){1-6}
\textit{Model} & \multicolumn{1}{c}{\texttt{GPT-4o}} & \multicolumn{1}{c}{\texttt{GPT-4o-mini}} & \multicolumn{1}{c}{\texttt{GPT-4o-mini}} & \multicolumn{1}{c}{\texttt{GPT-5-mini}} & - \\
\midrule
No Memory       & 0.485 & 0.673 & 0.841 & 0.836 & 0.709 \\
RAG             & 0.479 & 0.713 & 0.920 & 0.950  & 0.765 \\
Self-RAG        & 0.500 & 0.653 & 0.881 & 0.950 & 0.746 \\
Mem0            & 0.485 & 0.686 & 0.935 &  0.950 & 0.764 \\
MemP            & 0.494 & 0.720 & 0.928 & 0.921 & 0.766 \\
\midrule
\textit{MemRL (ours)}       & \textbf{0.508} & \textbf{0.746} & \textbf{0.942} & \textbf{0.979} & \textbf{0.794} \\
\bottomrule
\end{tabular}

\end{table*}


\subsection{Main Results}

\paragraph{Runtime Learning Results.}

As detailed in Table~\ref{tab:runtime_main_results}, \textsc{MemRL} demonstrates robust superiority across all domains, surpassing the strongest baseline (MemP) by an average of $+3.8\%$ in Cumulative Success Rate (CSR). The gains are most significant in exploration-intensive environments like ALFWorld and OS tasks (both $+6.2\%$), while maintaining a steady lead on the challenging HLE benchmark ($+3.6\%$). This confirms that our value-based mechanism, unlike MemP's heuristic retrieval, effectively filters noise to retain high-utility procedural patterns.

\paragraph{Transferring Results.}

We evaluate memory transferability by freezing the memory bank after training and testing on held-out sets.  As shown in Table~\ref{tab:transfer_results}, \textsc{MemRL} exhibits superior transferability, outperforming the strongest baseline (MemP) by an average of $+2.8\%$ in Success Rate. The advantage is particularly pronounced in complex environments like ALFWorld ($+5.8\%$) and OS tasks ($+2.6\%$). These margins validate that our \textbf{Two-Phase Retrieval} effectively filters low-value noise, retaining high-utility procedural patterns that generalize robustly to unseen scenarios.

\subsection{Ablations}
\subsubsection{Effectiveness of Runtime RL}

\begin{figure}[t]
    \centering

    \includegraphics[width=0.8\linewidth]{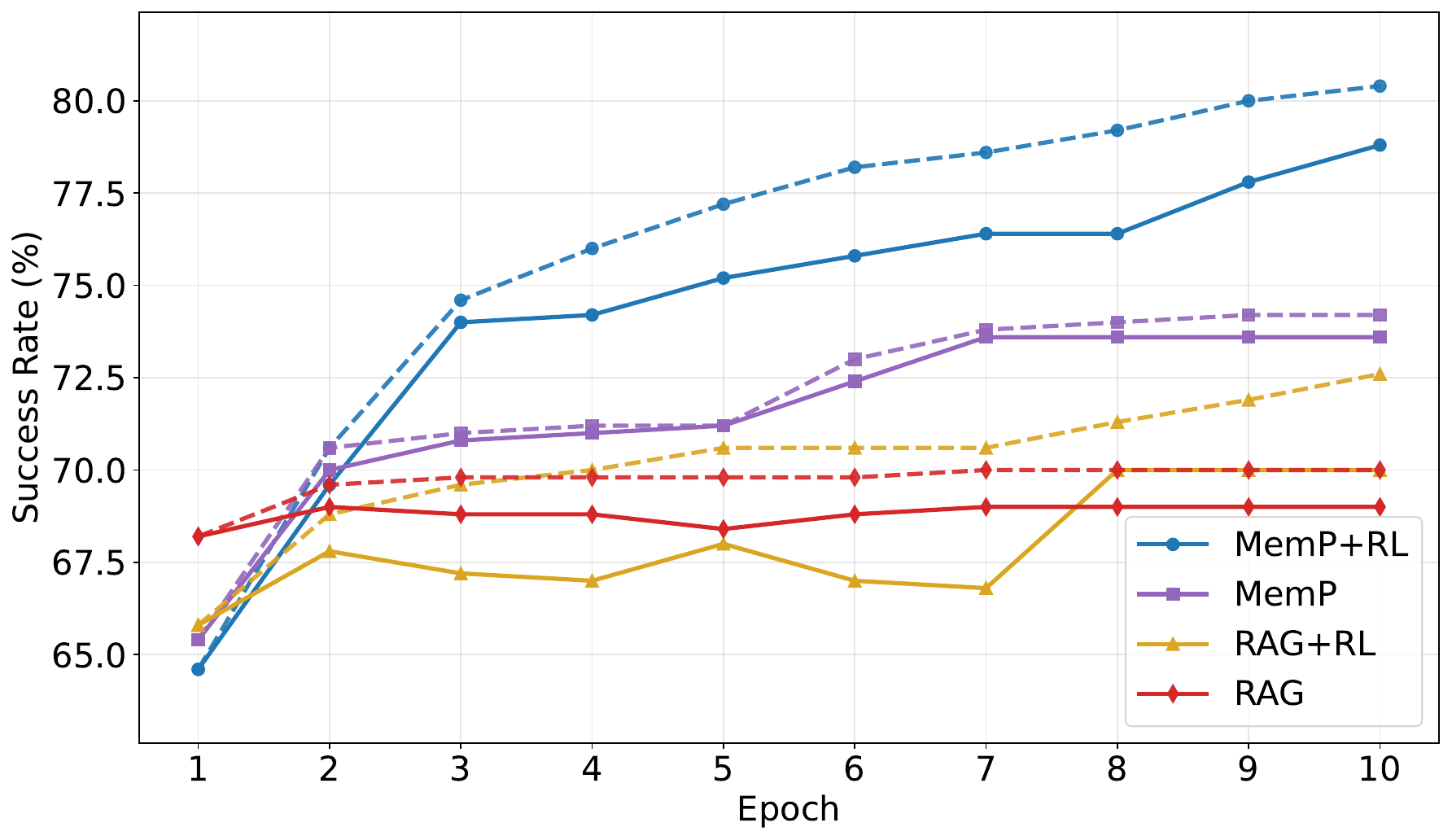}
    
    \caption{\textbf{OS Interaction Performance.} 
    Performance comparison of \textsc{MemRL} vs. baselines (MemP, RAG). \textbf{Solid lines} represent the Epoch Success Rate, while \textbf{dashed lines} indicate the Cumulative Success Rate (CSR). \textsc{MemRL} demonstrates superior stability and a widening performance gap in CSR.}
    \label{fig:os_performance_combined}
    \vskip -8pt
\end{figure}

To isolate the efficacy of runtime RL, we compare \textsc{MemRL} and its RAG-based variant against their non-RL counterparts (MemP and standard RAG) in the OS interaction environment. As shown in Figure~\ref{fig:os_performance_combined}, while initial performance is comparable, a clear divergence emerges as training progresses: \textsc{MemRL} achieves a smoother learning curve and superior stability.
Crucially, this advantage is most pronounced in the Cumulative Success Rate (dashed lines), where the monotonic widening gap indicates that the RL-driven value function effectively filters noisy memories and consolidates successful experiences.

\subsubsection{Impact of Q-Value Weighting}
\label{sec:ablation_lambda}

\begin{figure}[t]
    \centering
    \includegraphics[width=0.8\linewidth]{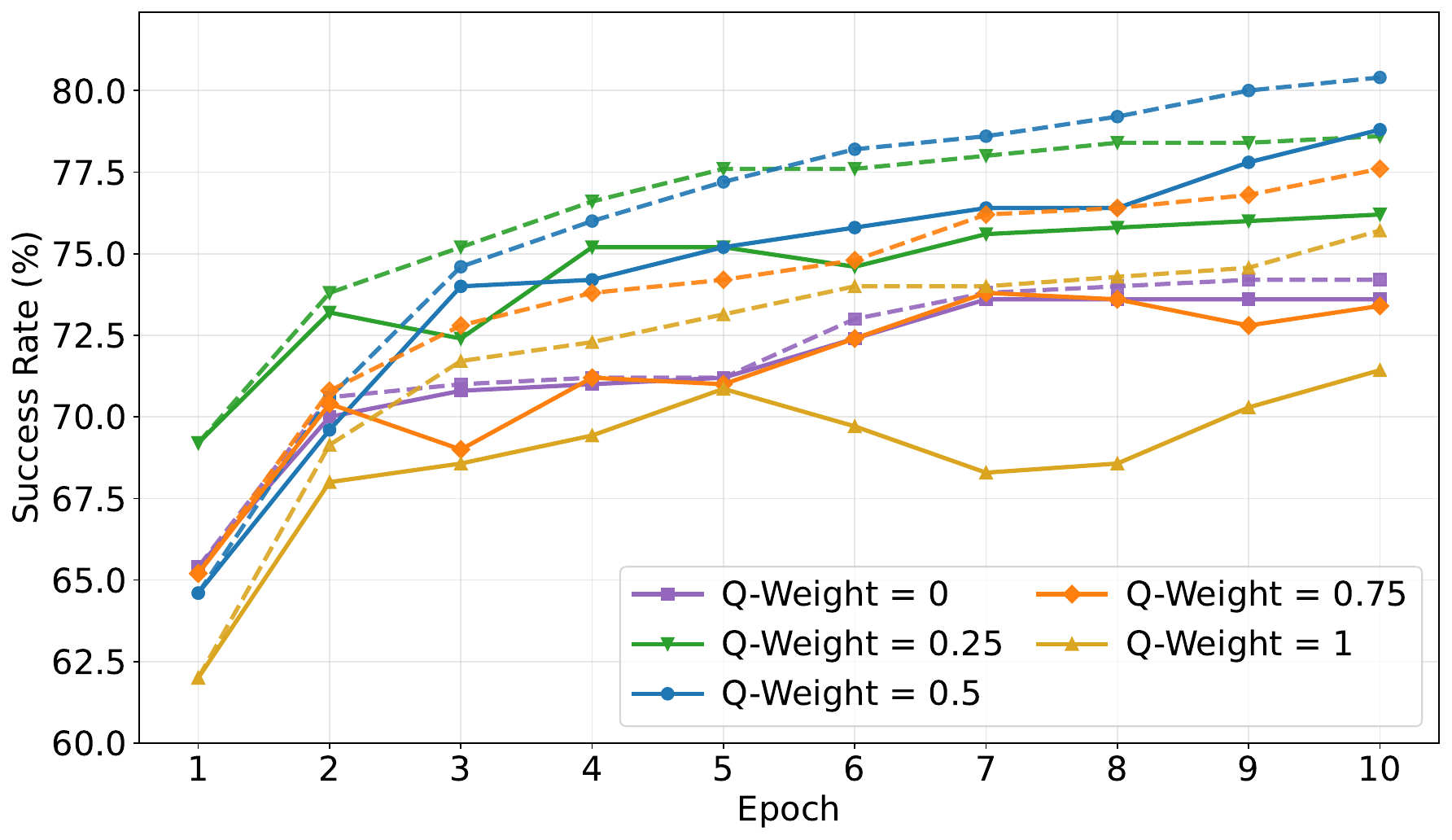}
    
    \caption{\textbf{Ablation on Q-value weighting factor $\lambda$.} 
    Performance comparison across $\lambda \in \{0, 0.25, 0.5, 0.75, 1\}$. \textbf{Solid lines} denote Epoch Success Rate, while \textbf{dashed lines} indicate Cumulative Success Rate (CSR). The balanced setting ($\lambda=0.5$) achieves the optimal trade-off between relevance and helpfulness.}
    \label{fig:q_weight_ablation}
    
    \vskip -0.15in
\end{figure}

To determine the optimal equilibrium between semantic grounding and value-based exploitation, we evaluate the Q-weighting factor $\lambda \in \{0, 0.25, 0.5, 0.75, 1\}$. As shown in Figure~\ref{fig:q_weight_ablation}, performance exhibits a clear concave trend peaking at $\lambda=0.5$. Deviating toward extremes degrades results: pure semantic retrieval ($\lambda=0$) plateaus due to an inability to filter functional distractors, while excessive RL weight ($\lambda \to 1$) induces volatility and context detachment. This confirms that $\lambda=0.5$ represents an effective balance, where semantic similarity guarantees content relevance and Q-value ensures its helpfulness.

\begin{table}[h]
\centering
\caption{\textbf{Ablation on Retrieval Scope.} We compare \textsc{MemRL} against an ablated version restricted to \textit{Single-Task Reflection}.}
\label{tab:ablation_scope}
\resizebox{\linewidth}{!}{%
\begin{tabular}{l c c c c c | c}
\toprule
\textbf{Setting} & \textbf{BCB} & \textbf{OS} & \textbf{DB} & \textbf{ALFWorld} & \textbf{HLE} & \textbf{Avg.} \\
\midrule
Single-Task Reflection & 0.614 & 0.714 & 0.938 & 0.930 & \textbf{0.610} & 0.761 \\
\textbf{\textsc{MemRL} (Cross-Task)} & \textbf{0.627} & \textbf{0.804} & \textbf{0.972} & \textbf{0.981} & 0.606 & \textbf{0.798} \\
\bottomrule
\end{tabular}%
}
\end{table}

\subsubsection{Ablation Analysis: Cross-Task vs. Single-Task Optimization}
\label{sec:ablation_scope}

To investigate the source of our performance gains, we conduct an ablation study in Table~\ref{tab:ablation_scope} by restricting the memory retrieval scope. We compare the full \textsc{MemRL} (\textit{Cross-Task Retrieval}) against an ablated setting that only utilizes feedback from the single task instance, conceptually equivalent to Reflexion \citep{shinn2023reflexion}. \textsc{MemRL} demonstrates superior performance in structured environments, particularly on \textbf{OS-Agent} ($+9.0\%$) and \textbf{ALFWorld} ($+5.1\%$). These benchmarks exhibit high intra-dataset similarity, allowing \textsc{MemRL} to effectively perform \textit{horizontal transfer}—retrieving and adapting successful policies from semantically similar historical tasks. While on the HLE benchmark, the single-task baseline ($0.610$) is tied with \textsc{MemRL} ($0.606$). We attribute this to the HLE dataset's low internal semantic similarity ($0.186$), as detailed in Appendix~\ref{app:sim-analysis}. This prevents effective cross-task generalization, forcing the agent to rely solely on single feedback.

\subsubsection{Sensitivity to Retrieval Size ($k_1$ and $k_2$).}

\begin{figure}[t]
    \centering
    \begin{subfigure}{\linewidth}
        \centering
        \includegraphics[width=0.8\linewidth, height=5.0cm, keepaspectratio]{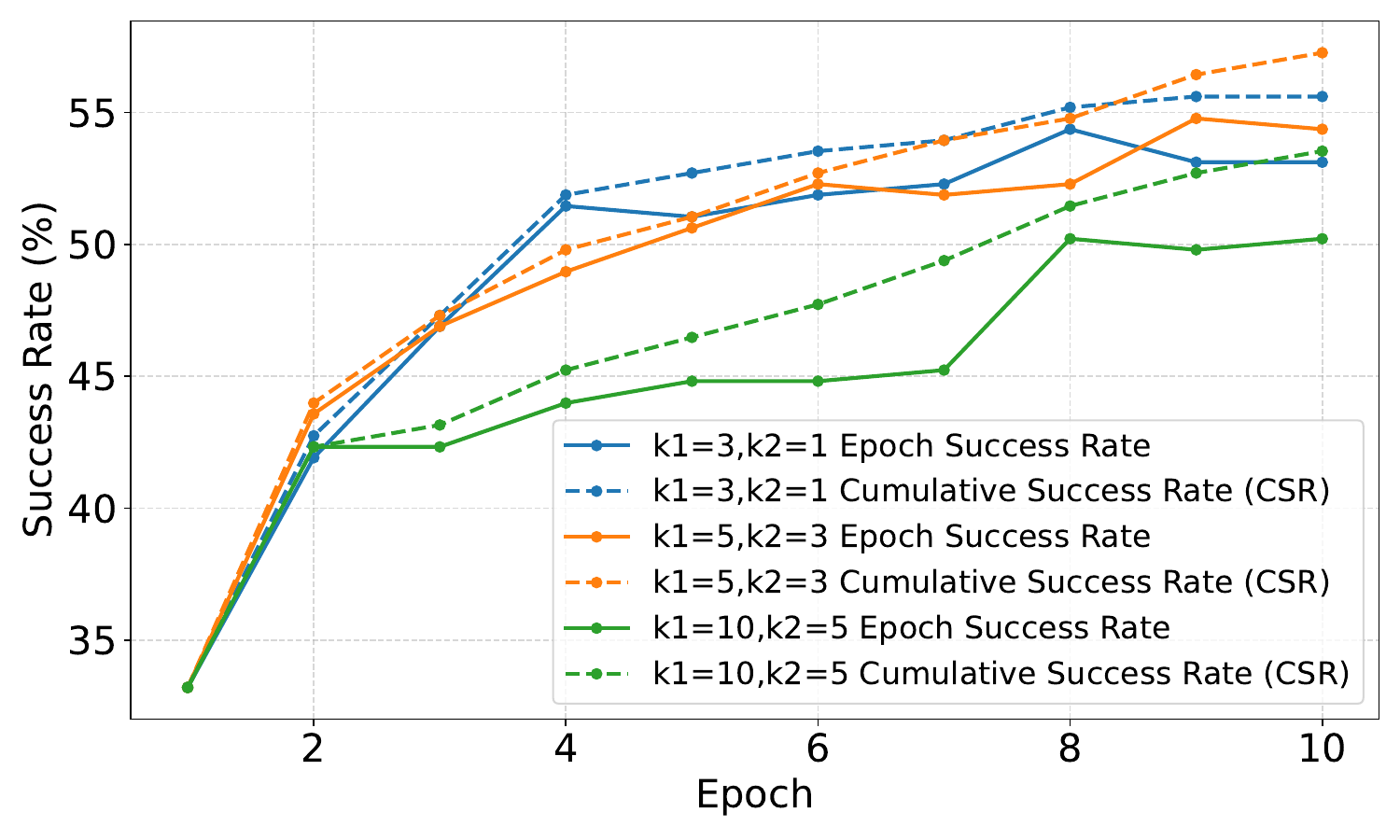}
    \end{subfigure}
    \caption{\textbf{Ablation on Retrieval Size ($k_1, k_2$).}
    Performance comparison on the HLE (CS/AI) subset across sparse ($3/1$), moderate ($5/3$), and dense ($10/5$) retrieval settings. The moderate setting achieves the optimal trade-off.}
    \label{fig:ablation_k}
    \vskip -0.15in
\end{figure}

To investigate the impact of retrieval bandwidth, we compare three memory density configurations on the HLE (CS/AI) subset benchmark: sparse ($k_1=3, k_2=1$), moderate ($k_1=5, k_2=3$), and dense ($k_1=10, k_2=5$). As shown in Figure \ref{fig:ablation_k}, performance follows an inverted-U trajectory, illustrating the trade-off between information sufficiency and context noise. The sparse setting limits performance due to insufficient guidance, whereas the dense setting degrades success rate by introducing distractions into the reasoning context. Consequently, the moderate configuration ($k_1=5, k_2=3$) achieves the best result, effectively maximizing the signal-to-noise ratio of the retrieved context.

\subsection{Discussion}

In this section, we delve deeper into the mechanisms driving \textsc{MemRL}'s performance, connecting empirical results to the challenge of balancing knowledge retention and adaptation.

\subsubsection{Predictive Power of the Q Critic}
    
    
    

\begin{figure}[t]
    \centering
    \begin{subfigure}[b]{0.8\linewidth}
        \centering
        \includegraphics[width=\linewidth]{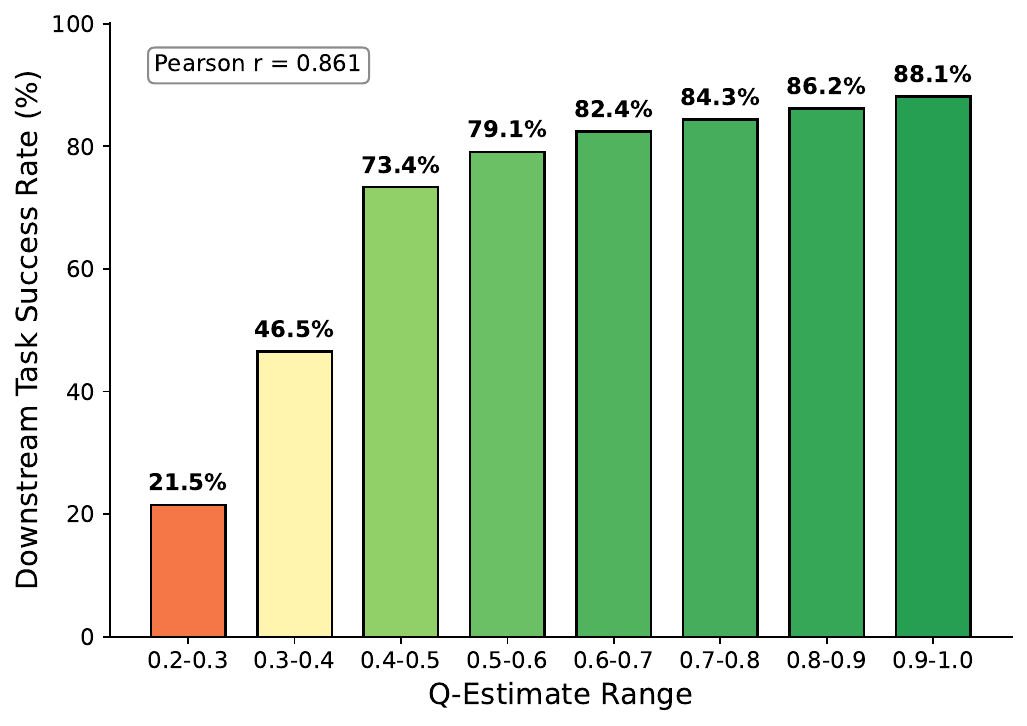}
        \vskip -4pt
        \caption{Success Rate vs. Q-Range}
        \label{fig:q_correlation}
    \end{subfigure}
    \hfill 
    \begin{subfigure}[b]{0.8\linewidth}
        \centering
        \includegraphics[width=\linewidth]{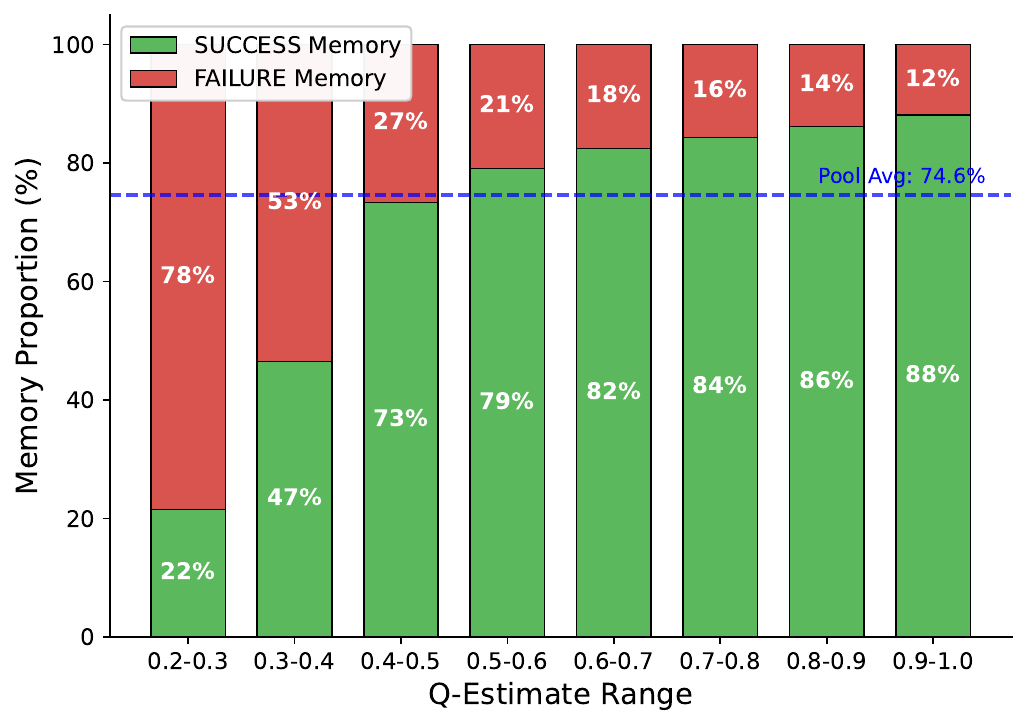}
        \vskip -4pt
        \caption{Memory Composition}
        \label{fig:q_composition}
    \end{subfigure}


    \caption{\textbf{Q-Value Analysis.} (a) Pearson $r=0.861$ confirms Critic's predictive power. (b) Failure memories ($\sim12\%$) in high Q-bins indicate latent strategic utility.}
    \label{fig:q_value_analysis}
    \vskip-10pt
    
\end{figure}







As shown in Figure~\ref{fig:q_correlation}, the learned Q-values exhibit a strong positive correlation (Pearson $r=0.861$) with empirical task success rates, rising from $21.5\%$ in the lowest-confidence bin to $88.1\%$ in the highest, which confirms the Critic's ability to effectively rank memories by success likelihood. 
Beyond simple ranking, memory composition analysis (Figure~\ref{fig:q_composition}) reveals that the agent retains a small fraction of ``failure'' memories ($\sim12\%$) even in high-Q bins ($0.9-1.0$), suggesting that Q-values capture \emph{utility beyond binary outcomes} by recognizing strategically useful near-misses. 
We further substantiate this with concrete case studies in Appendix~\ref{app:case}.
This indicates that the Critic prioritizes reusable guidance---including transferable procedural lessons from high-utility failures---rather than merely separating success from failure, thereby offering greater robustness than simple success-replay mechanisms.

\subsubsection{Stability of \textsc{MemRL}}
\label{sec:stable in mem rl exp}

\begin{figure}[t]
    \centering
    \begin{subfigure}{0.8\linewidth}
        \centering
        \includegraphics[width=\linewidth, height=5.5cm, keepaspectratio]{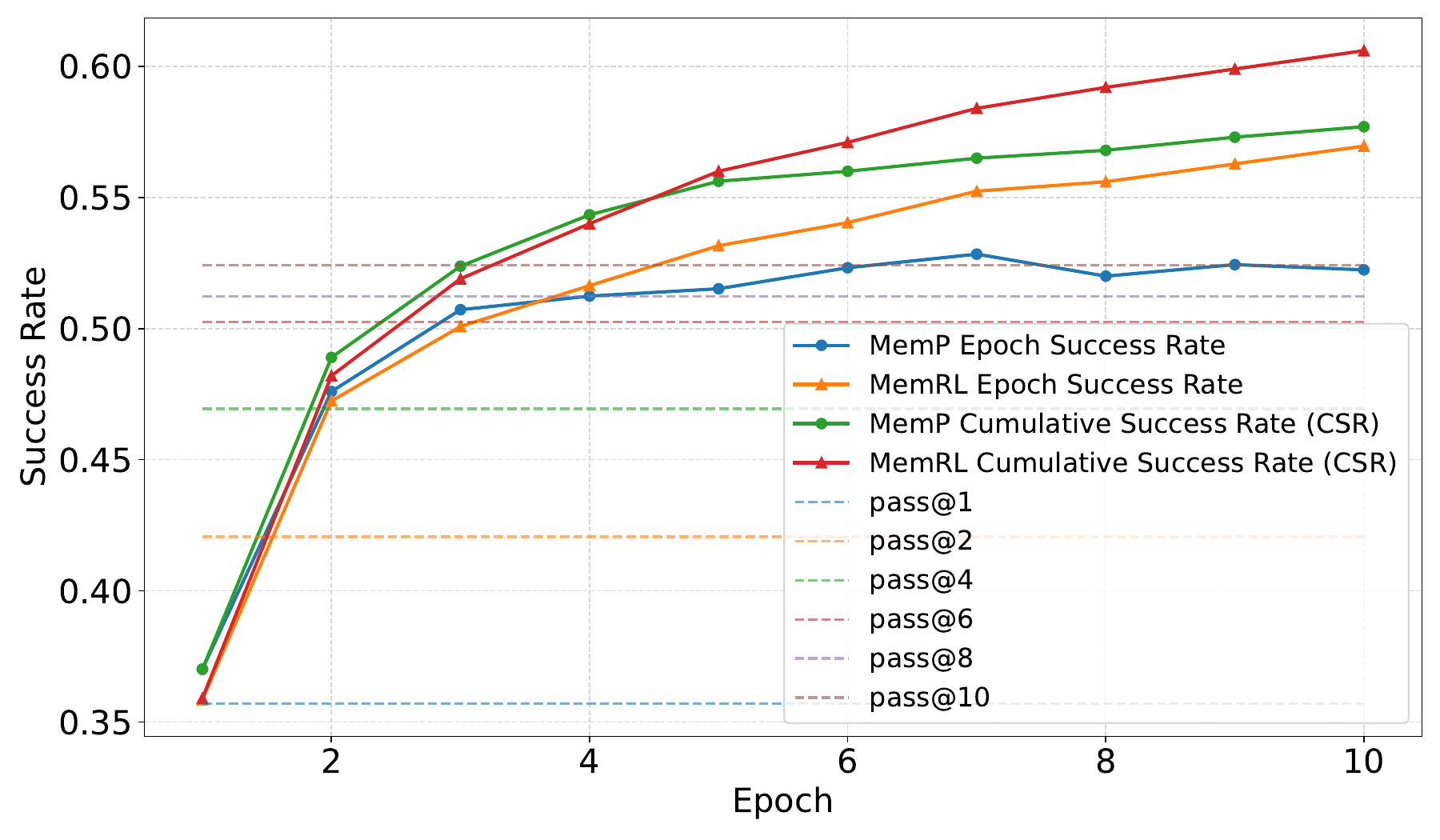}
    \end{subfigure}
    \vskip -4pt
    \caption{Epoch Success Rate and Cumulative Success Rate (CSR) of \textsc{MemRL} and MemP in HLE.}
    \label{fig:acc_gap}
    \vskip -6pt
\end{figure}

We analyze \textsc{MemRL} through the lens of the \textit{stability-plasticity dilemma}. The superior CSR (Table~\ref{tab:runtime_main_results}) confirms that \textsc{MemRL} effectively expands the solution space, enabling the agent to break through local optima. Furthermore, long-term dynamics (Figure~\ref{fig:acc_gap}) reveal a critical stability advantage: while heuristic methods like MemP suffer from catastrophic forgetting—evidenced by a widening gap between CSR and current Success Rate—\textsc{MemRL} maintains synchronized growth. This is theoretically guaranteed by our stability analysis in Section~\ref{sec:stability}, which constrains the policy to improve monotonically without drift.

We quantitatively validate these insights using the \textbf{Forgetting Rate}, defined as $FR = N_{\text{lost}} / N_{\text{fail}}$, where $N_{\text{lost}}$ denotes tasks transitioning from previous success to current failure and $N_{\text{fail}}$ is the total number of failures in the current epoch (see Figure~\ref{fig:transfer-forget} in Appendix~\ref{app:forgetting_dynamics} for the full trajectory). \textsc{MemRL} achieves the lowest mean forgetting rate ($0.041$), outperforming \textsc{MemP} ($0.051$). Additionally, ablation results demonstrate that removing z-score normalization and similarity gating causes the rate to spike to $0.073$. This confirms that strict filtering is essential to manage utility variance and ensure that self-evolution remains stable.
\subsubsection{Extended Analysis.} 
We conduct further investigations to characterize the underlying mechanisms and generalization of \textsc{MemRL}. Specifically, Appendix~\ref{app:analysis} analyzes \textsc{MemRL}'s role as a structural trajectory verifier and the correlation between task similarity and performance gains. Additionally, evaluations of advanced capabilities---including cross-model memory transferability and modular multi-task merging---are detailed in Appendix~\ref{app:advanced_exp}, demonstrating \textsc{MemRL}'s versatility and its capacity for modular capability expansion. We also analyze the cost and efficiency of \textsc{MemRL} in Appendix~\ref{app:efficiency_analysis}.

\section{Limitations and Conclusion}

\paragraph{Limitations and Future Work.}
While \textsc{MemRL} establishes a foundation for non-parametric evolution, its runtime dynamics reveal several promising avenues. (i) The current step-wise update, though fast, may introduce high-variance noise in long-horizon trajectories, inspiring us to explore multi-step updates or periodic memory consolidation. (ii) Credit-assignment ambiguity during utility updates, especially with multiple referenced experiences, raises the need for more precise attribution methods like Shapley methods~\citep{shapley1953value} or value decomposition in multi-agent reinforcement learning~\citep{rashid2020monotonic,sunehag2017value}. (iii) While \textsc{MemRL} improves with increasing task exposure, performance may drift toward reflection-like behavior when task similarity is low, highlighting the need for a sufficiently diverse yet relevant experience base; for industrial deployment, ensuring high task density and hierarchical abstraction may be crucial. Further detailed discussions on these and other challenges, including memory security, dedicated domains, and multi-agent memory sharing, are provided in Appendix~\ref{app:future_work}.

\paragraph{Conclusion.}
We proposed \textsc{MemRL}, a non-parametric approach reconciling the stability-plasticity dilemma by treating memory retrieval as a value-based decision process. Through the Intent-Experience-Utility triplet structure and Monte Carlo style updates, \textsc{MemRL} enables agents to self-evolve and differentiate high-utility strategies from semantic noise without weight updates. Extensive evaluations confirm \textsc{MemRL} significantly outperforms baselines in both runtime adaptation and generalization. In a future where static training data becomes scarce, the interactive experiences generated by agents throughout their life cycle will become a new, vital source of knowledge. We hope this paradigm paves the way for building stable, continuously learning agents that efficiently adapt from interaction.
\section*{Impact Statement}
This paper presents \textsc{MemRL}, a value-reinforced memory retrieval mechanism designed to enhance the long-term reasoning capabilities of LLM Agents. From a broader perspective, our work contributes to the development of more efficient and reliable autonomous systems. By optimizing the memory retrieval process, \textsc{MemRL} reduces the computational overhead of large-scale agent deployments, potentially lowering the environmental impact of AI infrastructure. Furthermore, as LLM agents become more integrated into daily workflows, research into robust memory mechanisms helps ensure these systems remain grounded and consistent in their actions. We do not foresee any immediate negative social consequences specific to this algorithmic advancement, though we acknowledge that all autonomous systems should be deployed with appropriate human oversight to mitigate broader risks associated with AI decision-making.

\bibliography{example_paper}
\bibliographystyle{icml2026}



\newpage
\onecolumn
\appendix
\raggedbottom

\section{Theoretical Analysis and Proofs}
\label{app:theoretical_analysis}
\subsection{Stability Analysis}
\label{app:stability}

We analyze the stability of \textsc{MemRL} from a reinforcement learning perspective, focusing on the convergence behavior of the utility estimates stored in memory. \textsc{MemRL} performs non-parametric runtime learning using a constant-step-size update. We show that under mild and realistic assumptions, the learned utility values converge in expectation to stable estimates of memory effectiveness, with bounded variance.

\paragraph{Setup.}
At each time step $t$, the agent observes an intent state $s_t$, retrieves a memory item $m_t \in \mathcal{M}_t$, generates an output $a_t$, and receives a scalar reward $r_t \in [-1,1]$ indicating task success or failure. The generation policy follows the decomposition defined in Eq.~\ref{eq:joint_policy}, where $\mu$ denotes the retrieval policy and $p_{\mathrm{LLM}}$ is a frozen inference policy.

For each retrieved memory, \textsc{MemRL} updates its utility using the exponential moving average rule as formulated in Eq.~\ref{eq:q_update}, with learning rate $\alpha \in (0,1]$. For clarity in this analysis, we consider a fixed state--memory pair $(s,m)$ and write $Q_t \equiv Q_t(s,m)$.

\paragraph{Stationary Reward Assumption.}
We analyze the learning process on a fixed dataset and posit two key conditions that ensure the stability of the environment:
\begin{enumerate}
    \item \textbf{Frozen Inference Policy.} The parameters of $p_{\mathrm{LLM}}(y | s,m)$ and the evaluator's criteria are fixed.
    \item \textbf{Fixed Task Distribution.} Tasks $s$ are drawn from a stationary distribution over a fixed dataset.
\end{enumerate}
These assumptions guarantee that the learning target is well-defined: the expected reward for any specific task-memory pair is time-invariant. Thus, we have:
\begin{equation}
\mathbb{E}[r_t | s_t = s,\; m_t = m] = \beta(s,m).
\label{eq:stationary_reward}
\end{equation}

where the expectation is taken over the stochastic rewards resulting from the Inference Policy $p_{LLM}$ distribution.

\paragraph{Expected Convergence of Utility Estimates.}
We now state the main stability result.

\begin{theorem}
\label{thm:ema_convergence}
Let $\{Q_t\}$ be updated according to the rule in Eq.~\ref{eq:q_update} with constant step size $\alpha \in (0,1]$. If Eq.~\ref{eq:stationary_reward} holds and the pair $(s,m)$ is updated infinitely often, then:
\begin{equation}
\label{eq:sm-stable}
\lim_{t \to \infty} \mathbb{E}[Q_t] = \mathbb{E}[r_t | s_t = s,\; m_t = m] = \beta(s,m).
\end{equation}
Moreover, the convergence rate is exponential \citep{sutton2018reinforcement}:
\begin{equation}
\mathbb{E}[Q_t] - \beta(s,m)
=
(1-\alpha)^t\big(Q_0 - \beta(s,m)\big).
\end{equation}
\end{theorem}

\paragraph{Proof.}
Define the estimation error $e_t \triangleq Q_t - \beta(s,m)$. Based on the update rule in Eq.~\ref{eq:q_update}, the error recurrence relation is:
\begin{equation*}
e_{t+1}
=
(1-\alpha)e_t
+
\alpha\big(r_t - \beta(s,m)\big).
\end{equation*}
Taking conditional expectation given the history $\mathcal{F}_t$ and using Eq.~\ref{eq:stationary_reward}, we obtain:
\begin{equation*}
\mathbb{E}[e_{t+1} | \mathcal{F}_t]
=
(1-\alpha)e_t.
\end{equation*}
Taking full expectation yields:
\begin{equation*}
\mathbb{E}[e_{t+1}] = (1-\alpha)\mathbb{E}[e_t].
\end{equation*}
Iterating the recursion gives $\mathbb{E}[e_t] = (1-\alpha)^t e_0$, which converges to zero as $t \to \infty$. \hfill $\square$

We provide the detailed derivation of the convergence proof in Appendix~\ref{app:proof_convergence}.

\paragraph{Bounded Variance and Stability.}
If the reward variance $\mathrm{Var}(r_t | s,m) < \infty$, then the variance of $Q_t$ remains bounded:
\begin{equation}
\limsup_{t \to \infty} \mathrm{Var}(Q_t)
\le
\frac{\alpha}{2-\alpha}\,\mathrm{Var}(r_t | s,m).
\end{equation}
Thus, constant-step-size updates do not induce unbounded oscillations; instead, they yield stable utility estimates that track expected memory effectiveness while filtering high-frequency noise. We explicitly derive the variance bounds to demonstrate the global stability of the estimator under task clustering in Appendix~\ref{app:bounded_variance}.

\paragraph{Global Stability via EM Convergence.}
The stability of the local estimate (Theorem~\ref{thm:ema_convergence}) extends to the global memory utility $Q(m)$. By the linearity of expectation, $Q(m)$ acts as a Monte Carlo integrator striving to converge to:

\begin{equation}
\label{eq:global_convergence}
\lim_{t \to \infty} \mathbb{E}[Q_t(m)] 
= \mathbb{E}[r | m] 
= \sum_{s \in \mathcal{S}(m)} \underbrace{\mathbb{E}[r | s, m]}_{\text{Stationary}} \underbrace{\Pr(s | m)}_{\text{Retrieve-Dependent}}.
\end{equation}

where $\mathcal{S}(m) \triangleq \{ s \in \mathcal{S} | \text{sim}(s, z_m) \ge \tau_A \}$ denotes the \textit{effective support set} for memory $m$, comprising all task intents $s$ sufficiently similar to the memory's intent embedding $z_m$ to satisfy the Phase-A retrieval criterion.

A theoretical challenge arises here: the weighting term $\Pr(s | m)$ is a latent variable governed by the retrieval policy $\mu(m| s; \mathcal{M})$, which itself shifts as $Q$-values evolve. To prove convergence despite this dependency, we analyze \textsc{MemRL} as a \textbf{Generalized Expectation-Maximization (GEM)} process~\citep{dempster1977maximum, neal1998view}.
From a variational perspective, the system performs coordinate ascent on a global objective function $\mathcal{J}(Q, \mu)$ (the variational lower bound of expected reward):
(i) \textbf{E-Step (Policy Improvement):} The Phase-B ranking updates the retrieval policy $\mu$ to align with current estimates, monotonically increasing $\mathcal{J}$ with respect to $\mu$;
(ii) \textbf{M-Step (Value Update):} The utility update (Eq.~\ref{eq:q_update}) increases $\mathcal{J}$ with respect to $Q$.
By the \textit{Monotonic Improvement Theorem}~\citep{neal1998view}, this alternating optimization guarantees that the system converges to a stationary point where the retrieve policy stabilizes ($\mu_{t+1} \approx \mu_t$). Consequently, the induced distribution $\Pr(s | m)$ becomes time-invariant, ensuring that Eq.~\ref{eq:global_convergence} holds and effectively preventing catastrophic forgetting by anchoring updates to a stable policy. More details can be found in Appendix~\ref{app:gem_analysis}.

\subsection{Proof of Theorem~\ref{thm:ema_convergence}: Convergence of EMA Estimation}
\label{app:proof_convergence}

We aim to prove that for a fixed task-memory pair $(s,m)$ with a stationary reward distribution, the Q-value estimate $Q_t(s,m)$ converges in expectation to the true mean reward $\beta(s,m)$.

\paragraph{Assumptions.}
\begin{enumerate}[leftmargin=*]
    \item \textbf{Stationary Reward.} The reward $r_t$ at step $t$ is drawn from a distribution induced by the stochastic action generation $a \sim p_{LLM}(a_t | s_t, m)$, with a constant mean $\beta(s,m) = \mathbb{E}[r_t | s, m]$ and finite variance $\sigma^2$.
    \item \textbf{Update Rule.} The utility is updated via the linear EMA rule with learning rate $\alpha \in (0,1)$:
    \begin{equation*}
        Q_{t+1} = (1-\alpha)Q_t + \alpha r_t.
    \end{equation*}
\end{enumerate}

\paragraph{Derivation of Error Dynamics.}
Let $e_t \triangleq Q_t - \beta(s,m)$ be the estimation error at time step $t$. Substituting $Q_t = e_t + \beta(s,m)$ into the update rule:

\begin{align}
    e_{t+1} + \beta(s,m) &= (1-\alpha)(e_t + \beta(s,m)) + \alpha r_t \notag \\
    e_{t+1} &= (1-\alpha)e_t + (1-\alpha)\beta(s,m) + \alpha r_t - \beta(s,m) \notag \\
    e_{t+1} &= (1-\alpha)e_t + \beta(s,m) - \alpha\beta(s,m) - \beta(s,m) + \alpha r_t \notag \\
    e_{t+1} &= (1-\alpha)e_t + \alpha (r_t - \beta(s,m)). \label{eq:error_recurrence}
\end{align}

\paragraph{Convergence Analysis.}
We define $\mathcal{F}_t$ as the filtration (history) up to time $t$. Since the reward $r_t$ depends on the action $a_t$ sampled subsequently from the frozen LLM, taking the conditional expectation of Eq.~\ref{eq:error_recurrence} given $\mathcal{F}_t$:

\begin{equation*}
    \mathbb{E}[e_{t+1} | \mathcal{F}_t] = (1-\alpha)e_t + \alpha (\underbrace{\mathbb{E}[r_t | \mathcal{F}_t]}_{\beta(s,m)} - \beta(s,m)) = (1-\alpha)e_t.
\end{equation*}

By the Law of Iterated Expectations, taking the full expectation yields:
\begin{equation*}
    \mathbb{E}[e_{t+1}] = \mathbb{E}[\mathbb{E}[e_{t+1} | \mathcal{F}_t]] = (1-\alpha)\mathbb{E}[e_t].
\end{equation*}

Iterating this recurrence relation from $t=0$:
\begin{equation*}
    \mathbb{E}[e_t] = (1-\alpha)^t \mathbb{E}[e_0].
\end{equation*}

Since $0 < \alpha < 1$, we have $|1-\alpha| < 1$. Consequently:
\begin{equation}
    \lim_{t \to \infty} \mathbb{E}[e_t] = \mathbb{E}[e_0] \cdot \lim_{t \to \infty} (1-\alpha)^t = 0.
\end{equation}

This proves that the estimator is unbiased in the limit, i.e., $\lim_{t \to \infty} \mathbb{E}[Q_t] = \beta(s,m)$. \hfill $\square$

\subsection{Bounded Variance and Global Stability}
\label{app:bounded_variance}

In this section, we provide the formal derivation for the variance bound of the estimator $Q_t$. We explicitly derive the finite-time variance formula via recursive unrolling and prove its asymptotic convergence, demonstrating how Phase-A clustering contributes to global stability.

\paragraph{Derivation of the Variance Bound.}
Let $\sigma^2 \triangleq \mathrm{Var}(r_t | s, m)$ be the variance of the reward signal, assumed to be finite. The EMA update rule is given by:
\begin{equation*}
    Q_{t+1} = (1-\alpha)Q_t + \alpha r_t.
\end{equation*}
Since the reward $r_t$ (current noise) is statistically independent of the current estimate $Q_t$ (which is determined by history $\mathcal{F}_{t-1}$), the variance of the sum is the sum of the variances:
\begin{align*}
    \mathrm{Var}(Q_{t+1}) 
    &= \mathrm{Var}((1-\alpha)Q_t) + \mathrm{Var}(\alpha r_t) \\
    &= (1-\alpha)^2 \mathrm{Var}(Q_t) + \alpha^2 \sigma^2.
\end{align*}
Let $v_t \triangleq \mathrm{Var}(Q_t)$. We obtain a linear recurrence relation $v_{t+1} = (1-\alpha)^2 v_t + \alpha^2 \sigma^2$.

\paragraph{Recursive Unrolling.}
To solve for $v_t$, we expand the recurrence relation backward from step $t$:
\begin{align}
    v_t &= (1-\alpha)^2 v_{t-1} + \alpha^2 \sigma^2 \notag \\
    &= (1-\alpha)^2 \left[ (1-\alpha)^2 v_{t-2} + \alpha^2 \sigma^2 \right] + \alpha^2 \sigma^2 \notag \\
    &= (1-\alpha)^4 v_{t-2} + \alpha^2 \sigma^2 \left[ 1 + (1-\alpha)^2 \right] \notag \\
    &\quad \vdots \nonumber \\
    &= (1-\alpha)^{2t} v_0 + \alpha^2 \sigma^2 \sum_{k=0}^{t-1} \left( (1-\alpha)^2 \right)^k. \label{eq:variance_summation}
\end{align}
Eq.~\ref{eq:variance_summation} explicitly shows that the variance at time $t$ consists of two components: the decayed initial variance (first term) and the accumulated noise variance (second term).

\paragraph{Asymptotic Convergence.}
As $t \to \infty$, since the learning rate $\alpha \in (0,1)$, the term $(1-\alpha)^{2t}$ vanishes. The summation term is a geometric series $\sum_{k=0}^{\infty} r^k = \frac{1}{1-r}$ with ratio $r = (1-\alpha)^2$. Thus:
\begin{equation*}
    \lim_{t \to \infty} v_t 
    = \alpha^2 \sigma^2 \cdot \frac{1}{1 - (1-\alpha)^2}.
\end{equation*}
Evaluating the denominator:
\begin{equation*}
    1 - (1-\alpha)^2 = 1 - (1 - 2\alpha + \alpha^2) = 2\alpha - \alpha^2 = \alpha(2-\alpha).
\end{equation*}
Substituting this back yields the tight variance bound:
\begin{equation}
    \limsup_{t \to \infty} \mathrm{Var}(Q_t) 
    = \frac{\alpha^2 \sigma^2}{\alpha(2-\alpha)} 
    = \frac{\alpha}{2-\alpha} \sigma^2.
\end{equation}

\paragraph{Connection to Phase-A Clustering.}
This result provides the theoretical justification for the stability of \textsc{MemRL}. While tasks within a memory cluster $\mathcal{S}(m) \triangleq \{ s | \text{sim}(s, z_m) > \tau_A \}$ may vary, the Smoothness Assumption implies their rewards are drawn from a distribution with bounded variance $\sigma^2_{\mathcal{S}(m)}$.
The derived bound $\frac{\alpha}{2-\alpha} \sigma^2_{\mathcal{S}(m)}$ guarantees that the memory utility $Q(m)$ will not diverge but will instead oscillate within a controlled range around the true expected utility. This mechanism effectively filters out high-frequency noise from diverse task instances while retaining the stable generalized value.

\subsection{Convergence via Variational Inference}
\label{app:gem_analysis}

In this section, we provide a theoretical foundation for \textsc{MemRL}, demonstrating that our retrieval strategy and update rules guarantee the convergence of value estimation.

\subsubsection{The Convergence Objective}
Our ultimate goal is to ensure that the estimated utility $Q(m)$ converges to the true expected return of memory $m$. This target value is defined as:
\begin{equation}
\label{eq:global_convergence2}
\lim_{t \to \infty} \mathbb{E}[Q_t(m)] 
= \mathbb{E}[r | m] 
= \sum_{s \in \mathcal{S}(m)} \underbrace{\mathbb{E}[r | s, m]}_{\text{Stationary}} \underbrace{\Pr(s | m)}_{\text{Retrieve-Dependent}}.
\end{equation}
The challenge lies in the term $\Pr(s | m)$—the probability that a specific state $s$ triggers the retrieval of $m$. This distribution depends on the retrieval policy $\mu_t(m|s)$, which itself evolves during training, creating a circular dependency that threatens stability.

\subsubsection{Variational Objective with Trust Region}
To resolve this, we formulate the problem as maximizing a \textbf{global variational objective} $\mathcal{J}(\mu, Q)$. This objective serves as a tractable lower bound for the global expected return defined in Eq.~\ref{eq:global_convergence2}, balanced by a semantic trust region:
\begin{equation}
\label{eq:variational_objective}
\mathcal{J}(\mu, Q) = \mathbb{E}_{s \sim \mathcal{D}} \left[ \underbrace{\sum_{m \in \mathcal{S}(s)} \mu(m|s) Q(s, m)}_{\text{Expected Utility } \approx \mathbb{E}[Q_t(m)]} - \frac{1}{\beta} \underbrace{D_{\text{KL}}\Big(\mu(\cdot|s) \big\| \pi_{\text{sim}}(\cdot|s)\Big)}_{\text{Semantic Trust Region}} \right]
\end{equation}
Here, the expectation is taken over the state distribution $\mathcal{D}$. The first term directly corresponds to the expected utility $\mathbb{E}[Q_t(m)]$ we aim to converge, while $\pi_{\text{sim}}$ represents the fixed semantic prior (derived from Phase-A). The KL-divergence term acts as a regularizer crucial for two reasons:
\begin{enumerate}[leftmargin=*]
    \item \textbf{Trust Region:} It constrains the policy to the support set $\mathcal{S}$, preventing the agent from retrieving high-Q but semantically irrelevant memories (out-of-distribution errors).
    \item \textbf{Regularization:} It stabilizes the learning dynamics during the ``cold start'' phase when Q-estimates are noisy.
\end{enumerate}

\subsubsection{Optimization via Generalized Expectation-Maximization (GEM)}
We treat the optimization of $\mathcal{J}$ as a GEM process, alternating between policy improvement and value evaluation:

\paragraph{E-Step (Policy Optimization).} 
We assume the utility estimates $Q(s, m)$ are fixed and seek the optimal retrieval policy $\mu^*$ that maximizes the global variational objective $\mathcal{J}(\mu, Q)$. Since the expectation is taken over the state distribution $\mathcal{D}$, we can maximize the objective for each state $s$ pointwise. The optimization problem for a specific state $s$ is:

\begin{equation}
\max_{\mu(\cdot|s)} \left[ \sum_{m \in \mathcal{S}(s)} \mu(m|s) Q(s, m) - \frac{1}{\beta} D_{\text{KL}}\Big(\mu(\cdot|s) \big\| \pi_{\text{sim}}(\cdot|s)\Big) \right]
\end{equation}
subject to the probability simplex constraint $\sum_{m \in \mathcal{S}(s)} \mu(m|s) = 1$.

Expanding the KL-divergence term, the objective function becomes:
\begin{equation}
\mathcal{L}(\mu) = \sum_{m \in \mathcal{S}(s)} \mu(m|s) \left( Q(s, m) + \frac{1}{\beta} \log \frac{\pi_{\text{sim}}(m|s)}{\mu(m|s)} \right).
\end{equation}

To enforce the normalization constraint, we introduce the Lagrange multiplier $\lambda$ and construct the Lagrangian:
\begin{equation*}
L(\mu, \lambda) = \sum_{m \in \mathcal{S}(s)} \mu(m|s) \left( Q(s, m) + \frac{1}{\beta} \log \frac{\pi_{\text{sim}}(m|s)}{\mu(m|s)} \right) + \lambda \left( 1 - \sum_{m \in \mathcal{S}(s)} \mu(m|s) \right).
\end{equation*}

Taking the derivative with respect to $\mu(m|s)$ and setting it to zero:
\begin{equation*}
\begin{aligned}
\frac{\partial L}{\partial \mu(m|s)} &= Q(s, m) + \frac{1}{\beta} \left( \log \frac{\pi_{\text{sim}}(m|s)}{\mu(m|s)} + \mu(m|s) \cdot \frac{\mu(m|s)}{\pi_{\text{sim}}(m|s)} \cdot \frac{-\pi_{\text{sim}}(m|s)}{\mu(m|s)^2} \right) - \lambda \\
&= Q(s, m) + \frac{1}{\beta} \left( \log \frac{\pi_{\text{sim}}(m|s)}{\mu(m|s)} - 1 \right) - \lambda = 0.
\end{aligned}
\end{equation*}

Rearranging terms to solve for $\mu(m|s)$:
\begin{equation*}
\begin{aligned}
\log \frac{\pi_{\text{sim}}(m|s)}{\mu(m|s)} &= \beta \lambda - \beta Q(s, m) + 1 \\
\frac{\mu(m|s)}{\pi_{\text{sim}}(m|s)} &= \exp\left( \beta Q(s, m) - (\beta \lambda + 1) \right) \\
\mu(m|s) &= \pi_{\text{sim}}(m|s) \exp(\beta Q(s, m)) \cdot \exp(-(\beta \lambda + 1)).
\end{aligned}
\end{equation*}

Since $\exp(-(\beta \lambda + 1))$ is independent of $m$, it acts as a normalization constant $1/Z(s)$. Thus, we recover the closed-form Boltzmann distribution used in our Phase-B retrieval:
\begin{equation*}
\mu^*(m|s) = \frac{\pi_{\text{sim}}(m|s) \exp(\beta Q(s, m))}{Z(s)} \propto \pi_{\text{sim}}(m|s) \exp(\beta Q(s, m)).
\end{equation*}
This derivation theoretically justifies our heuristic scoring function: the optimal retrieval policy naturally balances the semantic prior $\pi_{\text{sim}}$ and the learned utility $Q$.
By taking the logarithm, we recover the specific scoring function used in our \textbf{Phase-B Retrieval} (Eq.~\ref{eq:score}):
\begin{equation*}
\log \mu^*(m|s) \propto \underbrace{\log \pi_{\text{sim}}(m|s)}_{\approx \text{sim}(s, m)} + \beta Q_t(s, m)
\end{equation*}
This proves that our heuristic combination of similarity and Q-value is mathematically equivalent to the optimal policy under the variational objective.

\paragraph{M-Step (Policy Evaluation via Error Minimization).} 
While the E-step improves the policy based on current estimates, the M-step ensures these estimates are grounded in reality. Fixing the policy $\mu_{t+1}$, our goal is to align the variational parameter $Q$ with the true environmental returns. We formulate this as minimizing the Mean Squared Error (MSE) between the estimated utility and the observed reward target $y = r$ (in our Monte Carlo style modeling):
\begin{equation}
\min_Q \mathcal{L}(Q) = \mathbb{E}_{\tau \sim \mu_{t+1}} \left[ \frac{1}{2} \left( y - Q(s, m) \right)^2 \right]
\end{equation}
Minimizing this error is critical because it tightens the variational bound: it ensures that the expectation term $\mathbb{E}[Q]$ in the global objective $\mathcal{J}$ (Eq.~\ref{eq:variational_objective}) converges to the true expected return $\mathbb{E}[r]$.
The update rule used in our approach (Eq.~\ref{eq:q_update}) corresponds exactly to a Stochastic Gradient Descent (SGD) step on this objective:
\begin{equation*}
Q_{t+1}(s, m) \leftarrow Q_t(s, m) - \alpha \nabla_Q \mathcal{L}(Q) = Q_t(s, m) + \alpha (y - Q_t(s, m))
\end{equation*}
By iteratively minimizing $\mathcal{L}(Q)$, the M-step propagates the environmental feedback into the utility estimates, ensuring that the subsequent E-step optimization occurs on a reliable value landscape.

\subsubsection{Proof of Convergence}
By the \textit{Monotonic Improvement Theorem} of GEM~\citep{neal1998view}, the sequence $(\mu_t, Q_t)$ is guaranteed to converge to a stationary point $(\mu^*, Q^*)$.
At stationarity, the policy stabilizes ($\mu_{t+1} \approx \mu_t$), which implies that the inverse retrieval probability $\Pr(s|m)$ becomes \textbf{time-invariant}:
\begin{equation*}
\Pr(s|m) = \frac{\mu^*(m|s)\Pr(s)}{\sum_{s'} \mu^*(m|s')\Pr(s')}
\end{equation*}
Consequently, the ``Retrieve-Dependent'' term in Eq.~\ref{eq:global_convergence2} is anchored. With a fixed data distribution, the  $Q_t(m)$ converges to the unique fixed point:
\begin{equation}
\lim_{t \to \infty} Q_t(m) \to \mathbb{E}_{\mu^*}[r | m]
\end{equation}
Thus, our approach theoretically guarantees that the memory values converge to the true expected returns under the optimal retrieval policy.


\section{Extended Analysis and Insights}
\label{app:analysis}

\subsection{Detailed Analysis of Forgetting Dynamics}
\label{app:forgetting_dynamics}

In the main text, we reported the mean forgetting rate to quantify the stability of our method. Here, we provide a detailed visual analysis of how the forgetting rate evolves throughout the learning process on the HLE benchmark.

\begin{figure}[h] 
    \centering
    \includegraphics[width=0.5\linewidth]{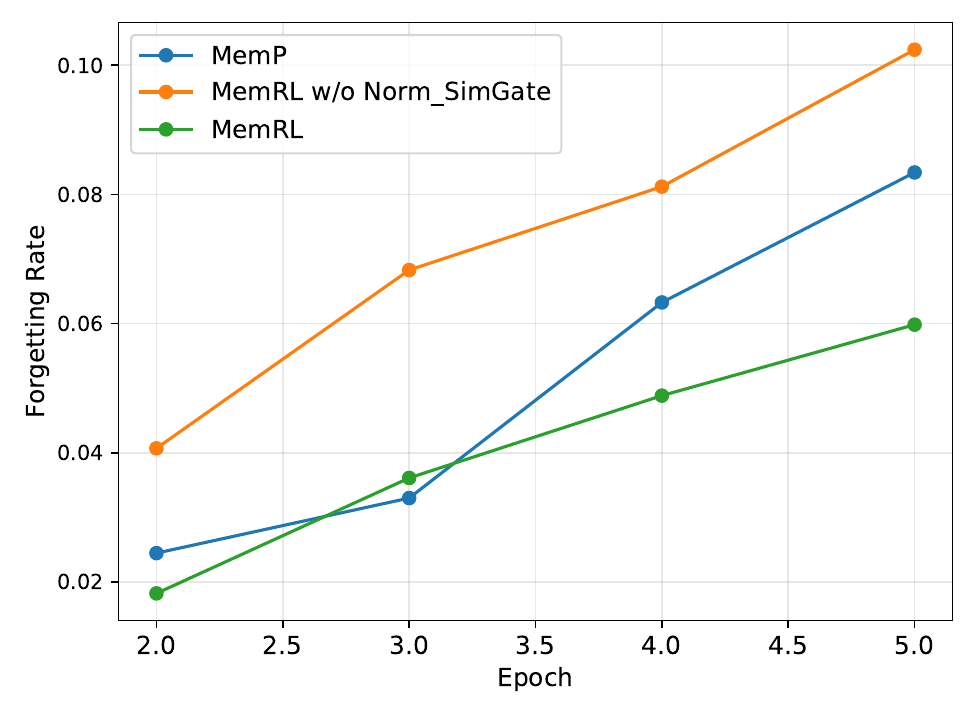}
    \caption{\textbf{Forgetting Rate Trajectory on HLE.} We compare the cumulative forgetting rate of the \textsc{MemRL} against the strong baseline MemP and an ablated version of \textsc{MemRL} (w/o Normalization \& Gating). A lower curve indicates better stability (less catastrophic forgetting).}
    \label{fig:transfer-forget}
\end{figure}

As illustrated in Figure~\ref{fig:transfer-forget}, \textsc{MemRL} demonstrates superior stability compared to MemP. Specifically:
\begin{itemize}
    \item \textbf{Stability vs. Plasticity:} While MemP shows a gradual upward trend in forgetting rate as the number of episodes increases, \textsc{MemRL} maintains a consistently low rate. This indicates that our dual-retrieval mechanism effectively balances the acquisition of new strategies without overwriting stable, high-utility memories.
    
    \item \textbf{Impact of Filtering:} The ablation curve (denoted as \textit{w/o Norm \& SimGate}) exhibits significant higher overall forgetting rate ($0.073$ mean). The visible spikes in the curve suggest that without z-score normalization and similarity gating, the agent frequently retrieves and reinforces ``noisy" strategies that work for specific instances but degrade general performance on previously mastered tasks.
\end{itemize}

\subsection{\textsc{MemRL} as a Trajectory Verifier.}

    
    
\begin{table}[h]  
    \centering
    \footnotesize
    \setlength{\tabcolsep}{1.5pt}

    \caption{\textbf{Impact of Task Structure.} Comparison of Cumulative Success Rate (CSR) gains. Multi-step tasks benefit significantly more from \textsc{MemRL}.}
    \label{tab:horizon_analysis}

    \begin{tabular}{@{}l l c c c@{}}
        \toprule
        \textbf{Benchmark} & \textbf{Interaction} & \textbf{MemP (\%)} & \textbf{\textsc{MemRL} (\%)} & \textbf{Gain (pp)} \\
        \midrule
        ALFWorld     & Multi-step  & 91.9 & \textbf{98.1} & \textbf{+6.2} \\
        OS Task      & Multi-step  & 74.2 & \textbf{80.4} & \textbf{+6.2}  \\
        HLE          & Single-step & 57.0 & 60.6 & +3.6 \\
        BigCodeBench & Single-step & 60.2 & 62.7 & +2.5 \\
        \bottomrule
    \end{tabular}
\vskip -5pt
\end{table}

Table~\ref{tab:horizon_analysis} reveals a correlation between task structural complexity and performance gain. The gains are most profound in multi-step sequential tasks (e.g., ALFWorld $+6.2\%$ Points(pp)) compared to single-turn tasks (e.g., BigCodeBench $+2.5\%$ pp).
In sequential tasks, a retrieved memory must be valid for the \textit{entire} trajectory. Standard semantic retrieval often fetches memories that match the initial instruction but fail in later steps. By propagating the final reward backward to the memory utility $Q$, \textsc{MemRL} effectively learns to verify the \textit{whole trajectory}, filtering out brittle policies that look correct only on the surface.

This analysis indicates that \textsc{MemRL} transcends the role of a simple retrieval enhancer to function as a \textit{Trajectory Verifier}. Its value is maximized in tasks with complex temporal dependencies, where it learns to select memories that ensure the structural integrity of the entire interaction process.

\subsection{Impact of Task Similarity on Memory Efficacy}
\label{app:sim-analysis}
To understand the underlying conditions where \textsc{MemRL} thrives, we analyze the correlation between the intra-dataset semantic similarity ($\text{Sim}_{intra}$) and the absolute performance gain provided by our method ($\Delta = \text{Success Rate}_{\text{MemRL}} - \text{Success Rate}_{\text{NoMem}}$).
    \begin{figure}[H]
    \centering
    \includegraphics[width=0.4\linewidth]{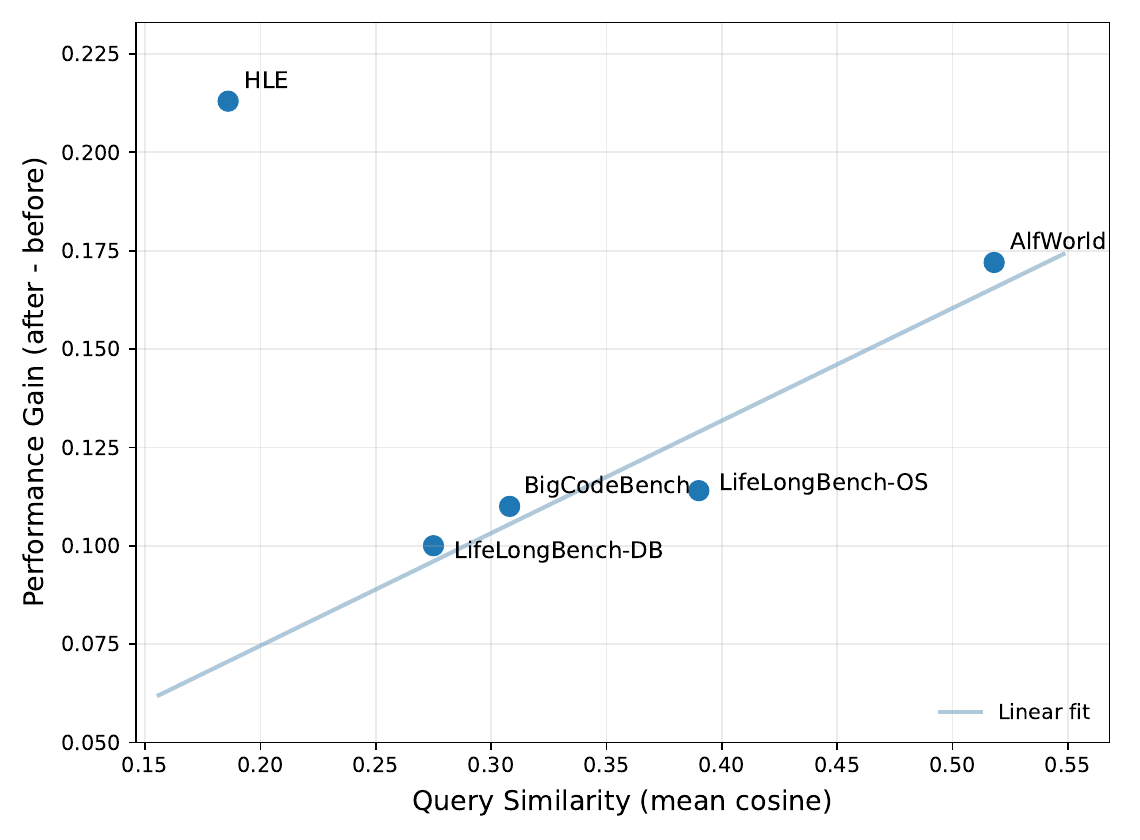}
    \caption{\textbf{Similarity-based Generalization.}}
    \label{fig:sim_gain}
\end{figure}
As illustrated in Figure~\ref{fig:sim_gain}, we analyze the correlation between intra-dataset semantic similarity and the absolute performance gain ($\Delta$) provided by \textsc{MemRL}. The linear regression trend reveals a general positive correlation: environments with higher structural repetition allow the agent to retrieve and reuse optimal policies more effectively. At the upper extreme, \textbf{ALFWorld} (similarity $0.518$) acts as a strong anchor point for this trend, exhibiting the highest repetition and a corresponding maximum performance boost ($\Delta=+0.172$). This confirms that for highly repetitive procedural tasks, memory serves as an effective shortcut to optimal trajectories. Following the regression line, benchmarks with moderate similarity—such as \textbf{Lifelong-OS} ($0.390$) and \textbf{BigCodeBench} ($0.308$)—cluster in the middle region, showing steady improvements ($\Delta \approx +0.10 \sim +0.11$) where the agent successfully generalizes coding patterns or OS commands across related instructions.  

\noindent \textbf{The HLE Anomaly: Generalization vs. Memorization.} 

HLE presents a unique outlier. Despite having the lowest similarity ($0.186$) due to its diverse, multi-disciplinary nature, it exhibits a surprisingly high runtime gain ($0.357 \rightarrow 0.570, \Delta=+0.213$).   
This gain operates on a different mechanism than ALFWorld. In high-similarity benchmarks, \textsc{MemRL} succeeds via \textit{Positive Transfer}—generalizing shared patterns to new instances. In contrast, the gain in HLE stems from \textit{Runtime Memorization}. Since HLE questions are distinct and domain-specific, the agent relies on the Runtime Learning phase to ``memorize" specific solutions to difficult problems through repeated exposure. This distinction highlights \textsc{MemRL}'s versatility: it supports both \textit{pattern generalization} in structured domains and \textit{specific knowledge acquisition} in diverse domains.  

\section{Advanced Capabilities: Transfer and Merging}
\label{app:advanced_exp}
\subsection{Cross-Model Memory Transferability}
\label{app:memory_transfer}

We investigate whether the procedural knowledge captured by \textsc{MemRL} is specific to the training policy or if it generalizes across different architectures. To test this, we take a memory bank fully trained for 10 epochs on the HLE benchmark using our strongest agent, \texttt{Gemini-3-pro}, and directly transfer it—without any fine-tuning—to three distinct inference models: \texttt{Qwen3-235B}, \texttt{GPT-5.2(High)}, and \texttt{Gemini-3-flash}.

As summarized in Table~\ref{tab:transfer_models}, the transferred memory yields substantial zero-shot performance gains across all models. Notably, smaller or distilled models experience the largest relative improvements; for instance, \texttt{Qwen3-235B} improves by over $3\times$ ($0.150 \rightarrow 0.531$) and \texttt{Gemini-3-flash} nearly doubles its success rate ($0.347 \rightarrow 0.583$). Even \texttt{GPT-5.2(High)}, a highly capable reasoning model, sees a significant boost ($0.354 \rightarrow 0.571$). 

These results suggest that \textsc{MemRL} captures \textit{model-agnostic} problem-solving patterns—such as efficient code skeletons and reasoning templates—rather than model-specific artifacts. This effectively allows the memory bank to function as a portable knowledge base, enabling weaker models to ``inherit" the capabilities of a stronger teacher model through simple retrieval.

\begin{table}[h]
\centering
\caption{\textbf{Cross-Model Transfer Results on HLE.} Performance of various agents using a frozen memory bank originally trained by \texttt{Gemini-3-pro}. $\Delta$ denotes the absolute improvement over the base model.}
\label{tab:transfer_models}
\resizebox{0.6\linewidth}{!}{%
\begin{tabular}{l c c c}
\toprule
\textbf{Inference Model} & \textbf{Base Score} & \textbf{Transfer Score} & \textbf{Gain ($\Delta$)} \\
\midrule
\texttt{Qwen3-235B} & 0.150 & 0.531 & \textbf{+0.381} \\
\texttt{Gemini-3-flash} & 0.347 & 0.583 & +0.236 \\
\texttt{GPT-5.2(High)} & 0.354 & 0.571 & +0.217 \\
\bottomrule
\end{tabular}%
}
\end{table}

\subsection{Multi-Task Memory Merging and Interference Analysis}
To evaluate the composability and robustness of \textsc{MemRL} in multi-task scenarios, we conducted a memory merging experiment. Specifically, we aggregated the finalized memory banks learned from the last epoch of two distinct domains within the Lifelong Agent Bench: Operating System Control ($M_{OS}$) and Database Management ($M_{DB}$). The agent was then evaluated on each respective benchmark using this unified, heterogeneous memory bank ($M_{Unified} = M_{OS} \cup M_{DB}$), without any further training or fine-tuning.

As presented in Table \ref{tab:memory_merge}, the results demonstrate that merging memories introduces negligible interference. The performance on the DB Task remains identical ($0.960$), while the OS Task exhibits only a marginal fluctuation ($0.788 \rightarrow 0.784$). This robustness is intrinsic to our \textbf{Two-Phase Retrieval} mechanism. Since the semantic spaces of OS commands and SQL queries are largely orthogonal, the Phase-A similarity filter effectively acts as a semantic gate, automatically excluding irrelevant cross-task memories before they enter the value-based ranking stage. This confirms that \textsc{MemRL} supports modular memory composition, allowing agents to scale capabilities by simply merging memory modules without suffering from negative transfer or catastrophic interference.

\begin{table}[H]
    \centering
    \small
    \caption{\textbf{Memory Merging Results.} Performance comparison between using task-specific individual memory banks versus a merged unified memory bank. The minimal delta confirms strong resistance to cross-task interference.}
    \label{tab:memory_merge}
    \setlength{\tabcolsep}{8pt}
    \begin{tabular}{l c c c}
        \toprule
        \textbf{Benchmark Task} & \textbf{Individual Memory} & \textbf{Merged Memory} & \textbf{Performance Delta} \\
        & (Original) & ($M_{OS} \cup M_{DB}$) & ($\Delta$) \\
        \midrule
        Lifelong-OS (OS Task) & 0.788 & 0.784 & -0.004 \\
        Lifelong-DB (DB Task) & 0.960 & 0.960 & \phantom{-}0.000 \\
        \bottomrule
    \end{tabular}
\end{table}

\section{Baseline and Benchmark Details}
\label{app:baseline_benchmark_details}

To ensure a rigorous evaluation, we compare \textsc{MemRL} against a diverse set of baselines ranging from simple sampling strategies to advanced agentic memory systems. All baselines are evaluated under a \textbf{unified frozen-backbone setting} to isolate the contribution of the memory and retrieval mechanisms.

\subsection{Baselines}
\label{app:baselines}
We categorize the baselines into three groups based on their interaction with memory and environment:

\paragraph{I. Test-Time Scaling Strategy}
\begin{itemize}
    \item \textbf{Pass@k:} This is a standard sampling-based baseline. It generates $k$ independent candidate solutions for a given query and selects the best one based on the benchmark's verifier (if available) or reports the success rate if at least one candidate passes. This baseline serves as a measure of the inherent capability of the frozen LLM without any memory persistence.
\end{itemize}

\paragraph{II. Retrieval-Augmented Generation (RAG) Approaches}
\begin{itemize}
    \item \textbf{RAG \citep{lewis2020retrieval}:} Represents the standard semantic retrieval paradigm. It utilizes an embedding model to encode the current query and retrieves the top-$k$ most semantically similar past experiences (or documents) from the external memory. These retrieved contexts are then prepended to the prompt to guide the LLM's generation.
    
    \item \textbf{Self-RAG \citep{asai2023selfrag}:} An advanced RAG variant that incorporates a self-critique mechanism. Unlike standard RAG, Self-RAG performs selective retrieval and uses a critique model (or self-prompting) to verify the relevance and factual correctness of the retrieved content before integrating it into the generation process.
\end{itemize}

\paragraph{III. Agentic Memory Systems}
\begin{itemize}
    \item \textbf{Mem0 \citep{chhikara2025mem0}:} A recently proposed memory layer for LLMs that manages memory through structured operations. It employs specific APIs for adding, retrieving, and updating memory, aiming to maintain a personalized and persistent context across interactions.
    
    \item \textbf{MemP \citep{fang2025memp}:} A framework focused on procedural memory. It distills past successful trajectories into reusable, procedure-like memory entries. MemP maintains a memory repository using a build-retrieve-update cycle, allowing the agent to recall high-level plans rather than raw trajectory data.
\end{itemize}

\subsection{Benchmark Datasets}
\label{app:benchmark}
We evaluate performance across four benchmarks selected to cover diverse domains: code generation, OS interactions, embodied decision-making, and multidisciplinary reasoning.

\begin{itemize}
    \item \textbf{BigCodeBench \citep{zhuo2025bigcode}:} A challenging benchmark for library-oriented code generation that requires agents to implement complex functionalities using diverse third-party libraries. Unlike traditional benchmarks focused on algorithmic snippets, BigCodeBench emphasizes practical software engineering capabilities. We evaluate on the \textbf{BigCodeBench-Instruct (Full)} split, which tasks the agent with synthesizing complete functional code from natural language instructions across the full range of difficulty levels.

    \item \textbf{Lifelong Agent Bench \citep{zheng2025lifelong}:} Designed to evaluate agents in a continuous learning setting involving Operating System (OS) and Database (DB) interactions. It tests the agent's capacity to adapt to new tools and commands over a long horizon without forgetting previous skills.
    
    \item \textbf{ALFWorld \citep{shridhar2021alfworld}:} An embodied navigation and manipulation benchmark. It requires the agent to solve textual logic puzzles within a simulated household environment (e.g., ``put a clean apple in the fridge"). This tests the agent's ability to learn and retrieve multi-step plans.
    
    \item \textbf{Humanity's Last Exam (HLE) \citep{phan2025hle}:} A rigorous multidisciplinary reasoning benchmark featuring hard problems from mathematics, humanities, and sciences. It serves as a stress test for the agent's general reasoning capability and its ability to retrieve relevant knowledge for disparate tasks.
\end{itemize}

\section{Implementation and Reproducibility Details}
\label{app:impl}

To facilitate reproducibility, we provide the exact model versions, hyperparameter settings, and environmental configurations used in our experiments.

\subsection{Model Specifications}
All LLM reasoning and generation tasks were performed using the models listed in Table \ref{tab:model_specs}. We used the official APIs with a fixed temperature to ensure deterministic evaluation where possible.

\begin{table}[H]
    \centering
    \caption{\textbf{Model and API Configurations.}}
    \label{tab:model_specs}
    \begin{tabular}{l l l}
        \toprule
        \textbf{Component} & \textbf{Configuration / Version} & \textbf{Notes} \\
        \midrule
        \textbf{Backbone LLM} & \texttt{GPT-4o} & Used for BigCodeBench \\
                              & \texttt{GPT-4o-mini} & Used for Lifelong Bench \\
                              & \texttt{GPT-5-mini} & Used for ALFWorld \\
                              & \texttt{Gemini-3-pro} & Used for HLE \\
        \midrule
        \textbf{Embedding Model} & \texttt{Text-Embedding-3-Large} & Used for Intent and Query encoding \\
        \midrule
        \textbf{Generation Params} & Temperature $= 0.0$ & General (Greedy decoding) \\
                                   & Temperature $= 0.6$ & Specific for HLE (\texttt{gemini-3-pro}) \\
                                   & Top-p $= 1.0$ & Default \\
        \bottomrule
    \end{tabular}
\end{table}

\subsection{Hyperparameter Settings}

Table \ref{tab:hyperparams} details the specific hyperparameters used for \textsc{MemRL} and the baselines. The similarity threshold $\delta$ is adaptive to the dataset density; specifically, we determine $\delta$ by calculating the pairwise cosine similarity distribution of task descriptions within each benchmark and selecting the threshold at the top 20\% quantile. This ensures that only the most relevant historical experiences are considered for retrieval.
\begin{table}[H]
    \centering
    \caption{\textbf{Hyperparameter Settings across Benchmarks.}}
    \label{tab:hyperparams}
    \resizebox{0.85\linewidth}{!}{%
    \begin{tabular}{l l c c c c c}
        \toprule
        & & \multicolumn{5}{c}{\textbf{Benchmark Setting}} \\
        \cmidrule(lr){3-7}
        \textbf{Parameter} & \textbf{Description} & \textbf{BigCodeBench} & \textbf{Lifelong Bench (OS)} & \textbf{Lifelong Bench (DB)} & \textbf{ALFWorld} & \textbf{HLE} \\
        \midrule
        \multicolumn{7}{l}{\textit{\textsc{MemRL} (Ours)}} \\
        $\alpha$ & Learning Rate (Eq.~\ref{eq:q_update}) & 0.3 & 0.3 & 0.3 & 0.3 & 0.3 \\
        $\lambda$ & Q-Weight Balance (Eq.~\ref{eq:score}) & 0.5 & 0.5 & 0.5 & 0.5 & 0.5 \\
        $\delta$ & Similarity Threshold & 0.38 & 0.50 & 0.37 & 0.62 & 0.25 \\
        $k_1$ & Phase-A Recall Size & 10 & 10 & 10 & 5 & 5 \\
        $k_2$ & Phase-B Select Size & 5 & 5 & 5 & 3 & 3 \\
        $Q_{init}$ & Initial Q-value & 0.0 & 0.0 & 0.0 & 0.0 & 0.0 \\
        \midrule
        \multicolumn{7}{l}{\textit{Baselines}} \\
        $k_{RAG}$ & Retrieval Top-k & 5 & 5 & 5 & 3 & 3 \\
        $k_{SelfRAG}$ & Retrieval Top-k & 5 & 5 & 5 & 3 & 3 \\
        $k_{Mem0}$ & Retrieval Top-k & 5 & 5 & 5 & 3 & 3 \\
        $k_{MemP}$ & Retrieval Top-k & 5 & 5 & 5 & 3 & 3 \\
        \bottomrule
    \end{tabular}%
    }
\end{table}
\subsection{Data Partitioning}
To evaluate the effectiveness of \textsc{MemRL}, we categorize our experiments into \textit{Runtime Learning} and \textit{Transfer Learning} settings. Table \ref{tab:data_split} summarizes the dataset sizes and partitioning strategies used for each benchmark. 

\begin{table}[H]
    \centering
    \footnotesize 
    \caption{\textbf{Summary of Data Partitioning and Dataset Sizes.}}
    \label{tab:data_split}
    \setlength{\tabcolsep}{8pt} 
    \begin{tabular}{l c c l}
        \toprule
        \textbf{Benchmark} & \textbf{Runtime Learning} & \textbf{Transfer Learning} & \textbf{Split / Note} \\
        \midrule
        HLE & 2,500 tasks  & -- & Full set (Runtime only) \\
        Lifelong Agent (OS) & 500 tasks & 500 tasks & 7:3 Split (Seed 42) \\
        Lifelong Agent (DB) & 500 tasks & 500 tasks & 7:3 Split (Seed 42) \\
        BigCodeBench-I (Full) & 1,140 tasks & 1,140 tasks & 7:3 Split (Seed 42) \\
        ALFWorld & 3,553 tasks & 140 tasks & \texttt{valid\_seen} (Novel instances)$^\dagger$ \\
        \bottomrule
    \end{tabular}
    \begin{flushleft}
    \scriptsize $^\dagger$ For ALFWorld Transfer Learning, the agent uses the same 3,553 tasks as memory context but is evaluated on 140 novel instances of seen task types.
    \end{flushleft}
\end{table}

For benchmarks utilizing random splits (\texttt{OS}, \texttt{DB}, and \texttt{BCB}), we use a fixed random seed of 42 to ensure reproducibility. In \texttt{ALFWorld}, the \textit{Transfer Learning} phase specifically tests generalization to new instances within known categories, ensuring the agent learns procedural patterns rather than specific trajectories.
\subsection{Model Selection and Performance Analysis}
\label{subsec:model_selection_regimes}

We select the backbone model for each benchmark to ensure a \emph{valid learning signal} relative to task complexity. Since \textbf{MemRL} relies on high-value trajectories to perform utility update over retrieved memories, extremely low initial competence can make the feedback effectively unusable. For instance, on challenging benchmarks such as HLE, weaker models may start at $\approx 4\%$ success rate (e.g., \texttt{GPT-4o-mini}), yielding too few successful trajectories for stable utility estimation; in this scenario, environmental feedback is dominated by noise, which can prevent meaningful learning and hinder convergence.

At the other extreme, using the strongest available models on simpler benchmarks can cause \emph{performance saturation} (ceiling effects), leaving little headroom to quantify the marginal gains attributable to the memory mechanism. Therefore, we match model capability to each task's difficulty to avoid both the \emph{no-signal situation} (insufficient successes) and the \emph{ceiling effect} (insufficient headroom). This design choice enables a more faithful evaluation of improvements that are intrinsic to \textbf{MemRL} rather than artifacts of an ill-posed performance condition.

Importantly, our evaluation spans backbones of different scales---from ``mini'' to ``pro'' tiers (Table~\ref{tab:model_specs})---to test whether \textbf{MemRL} remains effective across capability levels. The resulting performance trends support the \textbf{scale-invariant robustness} of \textbf{MemRL}. Moreover, the cross-model transfer results (Table~\ref{tab:transfer_models}) further indicate that the procedural knowledge captured in memory is portable across backbones, suggesting that the learned utility over memories generalizes beyond any single model's inherent strength.

To further explore the generality of \textbf{MemRL} across different model capabilities and task complexities, we also deployed the \texttt{GPT-4o-mini} model in the ALFWorld benchmark. This model is identical to the backbone used in the Lifelong Agent Bench, allowing for a direct comparison of its performance across environments with varying exploration intensities. As an exploration-intensive environment, ALFWorld poses significant demands on the base model's reasoning and planning capabilities. The Table~\ref{tab:runtime_learning_gpt4omini} and Table~\ref{tab:transfer_learning_gpt4omini} present the runtime learning and knowledge transfer abilities of \texttt{GPT-4o-mini}.

\begin{table}[htbp] 
\centering
\caption{Runtime Learning Performance of \texttt{GPT-4o-mini} Across Benchmarks}
\label{tab:runtime_learning_gpt4omini}
\begin{tabular}{lcccccccc}
\toprule
\multirow{2}{*}{\textbf{Method}} & \multicolumn{2}{c}{\textbf{Lifelong Agent (OS)}} & \multicolumn{2}{c}{\textbf{Lifelong Agent (DB)}} & \multicolumn{2}{c}{\textbf{ALFWorld}} & \multicolumn{2}{c}{\textbf{Average}} \\
\cmidrule(lr){2-3} \cmidrule(lr){4-5} \cmidrule(lr){6-7} \cmidrule(lr){8-9}
& \textbf{Last Epoch SR} & \textbf{CSR} & \textbf{Last Epoch SR} & \textbf{CSR} & \textbf{Last Epoch SR} & \textbf{CSR} & \textbf{Last Epoch SR} & \textbf{CSR} \\
\midrule
No Memory & 0.674 & N/A & 0.860 & N/A & 0.278 & N/A & 0.604 & N/A \\ 
Pass@10   & N/A   & 0.756 & N/A   & 0.928 & N/A & 0.462 & N/A  & 0.715 \\ 
MemP      & 0.736 & 0.742 & \textbf{0.960} & 0.966 & 0.299 & 0.413 & 0.665 & 0.707 \\
\textsc{MemRL}     & \textbf{0.788} & \textbf{0.804} & \textbf{0.960} & \textbf{0.972} & \textbf{0.440} & \textbf{0.680} & \textbf{0.730} & \textbf{0.819} \\ 
\bottomrule
\end{tabular}
\end{table}

\begin{table}[htbp]
\centering
\caption{Transfer Learning Performance of \texttt{GPT-4o-mini} Across Benchmarks}
\label{tab:transfer_learning_gpt4omini}
\begin{tabular}{lcccc}
\toprule
\textbf{Method} & \textbf{Lifelong Agent (OS)} & \textbf{Lifelong Agent (DB)} & \textbf{ALFWorld} & \textbf{Average} \\
\midrule
No Memory & 0.673 & 0.841 & 0.314 & 0.609 \\
MemP      & 0.720 & 0.928 & 0.314 & 0.654 \\
\textsc{MemRL}     & \textbf{0.746} & \textbf{0.942} & \textbf{0.457} & \textbf{0.715} \\
\bottomrule
\end{tabular}
\end{table}

These results, using \texttt{GPT-4o-mini} as a consistent backbone, unequivocally demonstrate \textsc{MemRL}'s superior and generalized effectiveness across all benchmarks. In runtime learning (Table~\ref{tab:runtime_learning_gpt4omini}), \textsc{MemRL} achieves the highest average Last Epoch Success Rate and Cumulative Success Rate, showing significant gains: its average Last Epoch SR is $12.6\%$ points higher than ``No Memory", and its average CSR surpasses MemP by $11.2\%$ points. This advantage is particularly pronounced in the challenging ALFWorld environment. Similarly, in transfer learning (Table~\ref{tab:transfer_learning_gpt4omini}), \textsc{MemRL} secures the highest average Success Rate, $10.6\%$  points higher than ``No Memory" and $6.1\%$ points higher than MemP. These consistent improvements across tasks and learning paradigms confirm that \textsc{MemRL} effectively enhances the \texttt{GPT-4o-mini} backbone's decision-making capabilities, leveraging learned utility over memories for more robust performance.

\section{Cost and Efficiency Analysis}
\label{app:efficiency_analysis}

Real-world deployment of autonomous agents requires balancing performance gains with computational costs. In this section, we analyze the token consumption and runtime latency of \textsc{MemRL} compared to the strong baseline MemP on the compute-intensive \textbf{Humanity's Last Exam (HLE)} benchmark.

\subsection{Token Consumption}

Since \textsc{MemRL} operates as a non-parametric approach without gradient-based fine-tuning, the primary cost arises from LLM API calls. We compare the average token usage per question (Q) across the entire learning trajectory (10 epochs).


\textsc{MemRL}'s token consumption is comparable to that of MemP, as both methods utilize identical interaction loops (reasoning + summarization). On the HLE benchmark, the average total token consumption per question for \textsc{MemRL} is approximately \textbf{32K}. This similarity in token usage is because the complexity of \textsc{MemRL} lies in \textit{how} memories are retrieved and updated (via Q-values), not in \textit{how much} context is fed to the LLM. Therefore, the significant performance gains of \textsc{MemRL} reported in the main text are achieved without increasing the inference budget.
\subsection{Runtime Latency and Stability}

\begin{figure}[h]
    \centering
    \includegraphics[width=0.5\linewidth]{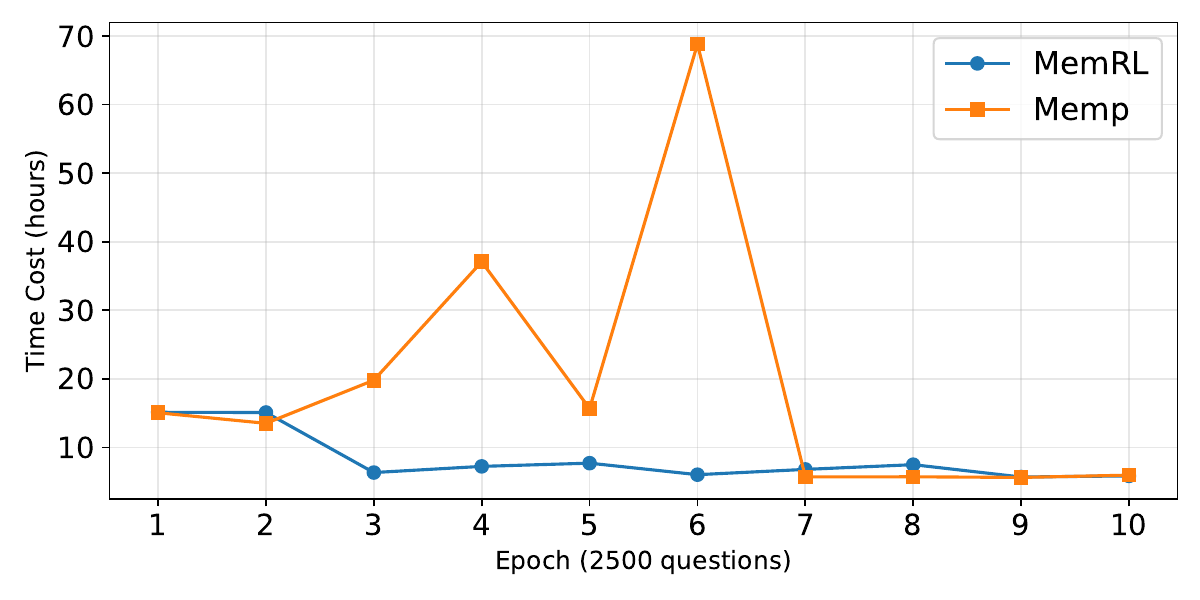}
    \caption{\textbf{Wall-Clock Time per Epoch on HLE.} The x-axis represents learning epochs (2,500 questions per epoch), and the y-axis represents the total time in hours. \textsc{MemRL} exhibits a stable runtime profile, confirming that the algorithmic overhead is negligible compared to system variances (e.g., network latency).}
    \label{fig:time_cost}
\end{figure}

A common concern with two-stage retrieval and reinforcement learning components is the potential for increased latency. We empirically validate the wall-clock time required to complete each epoch (2,500 questions) in Figure~\ref{fig:time_cost}.

\paragraph{Negligible Algorithmic Overhead.}
Figure~\ref{fig:time_cost} demonstrates that the runtime of \textsc{MemRL} is commensurate with, and often more stable than, that of MemP. The fluctuations observed (e.g., the spikes in the MemP curve at Epochs 4 and 6) are attributed to external factors such as API network latency and throughput variability, rather than algorithmic complexity.

Specifically, the additional components in \textsc{MemRL} introduce minimal computational cost:
\begin{itemize}
    \item \textbf{Dual-Stage Retrieval:} This involves basic vector dot-products and scalar score weighting, operating in milliseconds.
    \item \textbf{RL Update:} The Q-value update in a Monte Carlo style on scalars, which is an $O(1)$ operation.
\end{itemize}

Compared to the hours required for LLM generation over 2,500 questions, these millisecond-level operations are virtually imperceptible. The stability of the \textsc{MemRL} curve further suggests that our method does not introduce complex blocking operations that would exacerbate network-induced delays.

\section{Extended Limitations and Future Work}\label{app:future_work}

This appendix provides a more detailed discussion on the limitations of \textsc{MemRL} and outlines promising avenues for future research, building upon the foundations established in the main paper.

\subsection{Update Protocols and Memory Consolidation}
The current step-wise update protocol of \textsc{MemRL}, while enabling rapid adaptation, can introduce high-variance noise in long-horizon trajectories. This presents a challenge for stabilizing value estimation over extended sequences. A valuable future direction involves exploring more robust multi-step update mechanisms, which could offer slower but more stable learning. Furthermore, combining these with periodic consolidation of similar intentions and experiences within the memory bank could significantly improve the spatial efficiency of the memory, reducing redundancy and enhancing retrieval quality.

\subsection{Precise Credit Assignment in Multi-Memory Updates}
A significant challenge arises from the credit-assignment ambiguity encountered when multiple experiences from the memory are referenced and updated simultaneously. Determining the precise contribution of each referenced memory to the final outcome is complex and affects the efficiency of learning. Future research could investigate methods for more precise experience attribution, drawing inspiration from techniques like Shapley values~\citep{shapley1953value}, commonly used in cooperative game theory, or value decomposition methods prevalent in multi-agent reinforcement learning~\citep{rashid2020monotonic,sunehag2017value}. Such approaches could lead to more accurate updates and faster convergence.

\subsection{Task Similarity and Generalization}
\textsc{MemRL}'s effectiveness is observed to improve with the number of encountered tasks, aligning well with the characteristics of runtime learning. This is because the algorithm leverages past experiences, and a richer, more diverse set of similar experiences directly enhances its ability to retrieve relevant knowledge. However, when task similarity in the experience base is low, the method may inherently degrade into a less efficient reflection-like behavior, as direct experience transfer is limited. This is not a deficiency of the method but rather a characteristic of its reliance on learned utility from memory. For industrial deployment and real-world applications, this implies that maintaining a sufficiently high ``task similarity density" within the agent's operating environment or its collected memory is crucial. Strategies to achieve this could include active curriculum learning, environment design that promotes diverse yet related tasks, or techniques that facilitate hierarchical abstraction to enable generalization beyond direct, low-level experience matches. Enhancing retrieval mechanisms to proactively identify and adapt to novel task structures, or integrating hierarchical abstraction, are key for improving cross-task generalization.

\subsection{Memory Security and Robustness to Attack}
\textsc{MemRL} is sensitive to the quality of feedback, particularly vulnerable to ``reward hacking" if the verifier produces false positives. Incorrectly learned high Q-values from spurious feedback can quickly solidify and propagate erroneous behavioral patterns, leading to systemic failures. This highlights a critical challenge: memory security. A maliciously injected sample into the memory bank could rapidly diffuse pollution, potentially causing an intelligent agent to collapse. Future research must address the safety and trustworthiness of agent memories. Fortunately, the mutable nature of experience offers a silver lining: once contamination or attack is identified, polluted experiences can be swiftly pruned, allowing for recovery without disrupting the utility of previously learned valid experiences.

\subsection{Multi-Agent Collaboration and Shared Memory}
As many enterprises transition from single agents to multi-agent clusters (swarms), the question of ``shared memory" becomes important. Does the \textsc{MemRL} approach support knowledge sharing, allowing lessons learned and updated Q-values from one agent to immediately benefit others? We've explored scenarios where the memory, treated as a persistent resource, can indeed be shared among different models or agents (Appendix~\ref{app:memory_transfer}). Furthermore, adopting a multi-agent paradigm, akin to distributed parameter updates, could be highly beneficial. Each agent could accumulate experience and save it as memory during its operation, contributing to a large, collective memory pool. This would enable different agents to implicitly leverage each other's learned experiences, fostering a more collaborative and efficient learning ecosystem.

Beyond this collective pooling, a significant and promising frontier involves investigating the selective nature of such knowledge diffusion through the lens of transfer learning. As the number of agents increases, particularly beyond simple pairings, the challenge of "what to share" and "with whom to share" becomes a critical research direction. Future work could explore mechanisms that differentiate between universally applicable procedural insights and task-specific noise, ensuring that memory transfer remains contextually relevant and avoids the risks of negative transfer. Navigating these trade-offs between global collective intelligence and targeted knowledge distribution will be essential for building scalable, self-evolving swarms.
\subsection{Dedicated Domains and Hybrid Architectures}
While this work strongly advocates for freezing large language models (LLMs) to prevent catastrophic forgetting, a question arises regarding highly specialized domains where the base model may not comprehend foundational terminology. Is \textsc{MemRL} alone sufficient in such cases? We anticipate a future hybrid model where companies periodically fine-tune foundational models (e.g., annually) to update their core vocabulary and understanding, while simultaneously leveraging \textsc{MemRL} for daily behavioral adaptation and runtime learning. This approach requires the base model to possess a relatively high level of ``intelligence" to initially grasp and subsequently learn new terms through interaction and feedback.

\section{Case Study: High-Utility Failure Analysis (Near-Misses)}
\label{app:case}
This appendix section provides qualitative case studies of high-value near-miss memories mined from the 10-epoch OS-Interaction run.
CS denotes \emph{Case Study}. Each box below contains: origin task, retrieved reflection memory, a short explanation, and the target task where it was retrieved.

\tcbset{boxsep=1.0mm}
\newtcolorbox{CaseBox}[1]{%
  enhanced,
  breakable,
  colback=white,
  colframe=black!55,
  colbacktitle=black!6,
  coltitle=black,
  fonttitle=\bfseries,
  title={#1},
  boxrule=0.5pt,
  arc=2pt,
  left=6pt,right=6pt,top=6pt,bottom=6pt,
}
\newtcblisting{CodeBox}{%
  listing engine=listings,
  listing only,
  breakable,
  colback=black!1,
  colframe=black!35,
  boxrule=0.4pt,
  arc=2pt,
  left=6pt,right=6pt,top=4pt,bottom=4pt,
  listing options={
    basicstyle=\ttfamily\small,
    columns=fullflexible,
    keepspaces=true,
    breaklines=true,
    breakatwhitespace=false,
    showstringspaces=false
  },
}

\begin{CaseBox}{Case Study 1 (CS1)}
\textbf{Utility.} selected=19, success=19 (100.0\%).
\par\medskip
\textbf{Origin task.}
\begin{CodeBox}
Update the SSH daemon configuration using sed to set: Port=2222, PermitRootLogin=no, X11Forwarding=no, ClientAliveInterval=300, ClientAliveCountMax=2, MaxAuthTries=2, MaxSessions=2, PasswordAuthentication=no, AllowUsers='user1 user2', UseDNS=no, and Protocol=2.
\end{CodeBox}
\medskip
\textbf{Retrieved reflection memory.}
\begin{CodeBox}
- #1: [MEMORY TYPE] FAILURE_REFLECTION
[TASK]
Update the SSH daemon configuration using sed to set: Port=2222, PermitRootLogin=no, X11Forwarding=no, ClientAliveInterval=300, ClientAliveCountMax=2, MaxAuthTries=2, MaxSessions=2, PasswordAuthentication=no, AllowUsers='user1 user2', UseDNS=no, and Protocol=2.

[REFLECTION]
- ROOT CAUSE: The `sed` command did not modify the commented lines in the configuration file, which is why the new settings were not applied as expected.
- PATTERN TO AVOID: Avoid assuming that `sed` will replace commented lines without explicitly handling them.
- CORRECT APPROACH: Use `sed` to either uncomment the lines before replacing them or add the new configurations if they are commented out.
\end{CodeBox}
\par\medskip
\textbf{Explanation.} This reflection captures a subtle config-edit pitfall: naive sed replacement may miss commented defaults. The transferable lesson is to explicitly handle commented lines (uncomment-or-append), which generalizes across configuration-editing tasks.
\par\medskip
\textbf{Target task} (retrieval rank 1; run outcome: correct).
\begin{CodeBox}
Modify the SSH daemon configuration file at /etc/ssh/sshd_config to change the port to 2222, disable root login, set max sessions to 2, and set client alive interval to 300. Create a backup file before making changes.
\end{CodeBox}
\end{CaseBox}
\medskip

\begin{CaseBox}{Case Study 2 (CS2)}
\textbf{Utility.} selected=20, success=20 (100.0\%).
\par\medskip
\textbf{Origin task.}
\begin{CodeBox}
Replace 'DEBUG=True' with 'DEBUG=False' in /app/settings.cfg, append 'LOG_LEVEL=INFO' after the modified line, and create a backup file.
\end{CodeBox}
\medskip
\textbf{Retrieved reflection memory.}
\begin{CodeBox}
- #2: [MEMORY TYPE] FAILURE_REFLECTION
[TASK]
Replace 'DEBUG=True' with 'DEBUG=False' in /app/settings.cfg, append 'LOG_LEVEL=INFO' after the modified line, and create a backup file.

[REFLECTION]
- ROOT CAUSE: The `sed` command added the line `LOG_LEVEL=INFO` twice due to the way it was structured to append after the modified `DEBUG` line. 
- PATTERN TO AVOID: Avoid using multiple `-e` options in `sed` that can lead to unintended duplications when appending lines.
- CORRECT APPROACH: Instead, use a single `sed` command that ensures the line is only added once, or check for the existence of the line before appending it.
\end{CodeBox}
\par\medskip
\textbf{Explanation.} The originating attempt nearly completes the edit but fails due to a non-idempotent text transformation (duplicating an appended line). The memory teaches an idempotent editing pattern (single-pass edit or existence check), which transfers to many file-edit tasks.
\par\medskip
\textbf{Target task} (retrieval rank 2; run outcome: correct).
\begin{CodeBox}
Replace 'DEBUG=True' with 'DEBUG=False' in /app/settings.cfg, append 'LOG_LEVEL=INFO' after the modified line, and create a backup file.
\end{CodeBox}
\end{CaseBox}
\medskip

\begin{CaseBox}{Case Study 3 (CS3)}
\textbf{Utility.} selected=48, success=44 (91.7\%).
\par\medskip
\textbf{Origin task.}
\begin{CodeBox}
Create a user 'testuser' with a home directory, set password expiration policy (max 60 days, min 7 days, warning 7 days), add to 'testgroup', create '/home/testuser/private' with 770 permissions, create '/shared' directory accessible only by 'testgroup', add 'export PATH=$PATH:/custom' to .bashrc, and create '.hushlogin' file with 644 permissions.
\end{CodeBox}
\medskip
\textbf{Retrieved reflection memory.}
\begin{CodeBox}
- #4: [MEMORY TYPE] FAILURE_REFLECTION
[TASK]
Create a user 'testuser' with a home directory, set password expiration policy (max 60 days, min 7 days, warning 7 days), add to 'testgroup', create '/home/testuser/private' with 770 permissions, create '/shared' directory accessible only by 'testgroup', add 'export PATH=$PATH:/custom' to .bashrc, and create '.hushlogin' file with 644 permissions.

[REFLECTION]
- ROOT CAUSE: The output of the commands was empty, indicating successful execution without any visible confirmation or errors. 
- PATTERN TO AVOID: Relying solely on silent execution without feedback can lead to uncertainty about whether tasks were completed successfully. 
- CORRECT APPROACH: Always check the status of commands using conditional statements or output messages to confirm successful execution and provide clarity on the task's completion.
\end{CodeBox}
\par\medskip
\textbf{Explanation.} This memory highlights a common near-miss failure mode: assuming silent command output implies completion without verification. The transferable practice is to add explicit post-checks (exit codes, file/permission/ownership checks) after critical steps.
\par\medskip
\textbf{Target task} (retrieval rank 4; run outcome: correct).
\begin{CodeBox}
Create a user 'testuser' with a home directory, set password expiration policy (max 60 days, min 7 days, warning 7 days), add to 'testgroup', create '/home/testuser/private' with 770 permissions, create '/shared' directory accessible only by 'testgroup', add 'export PATH=$PATH:/custom' to .bashrc, and create '.hushlogin' file with 644 permissions.
\end{CodeBox}
\end{CaseBox}
\medskip

\begin{CaseBox}{Case Study 4 (CS4)}
\textbf{Utility.} selected=26, success=23 (88.5\%).
\par\medskip
\textbf{Origin task.}
\begin{CodeBox}
Run a sleep process in the background for 3 seconds, log its start and end times (in epoch format) to '/var/log/sleep.log', store the process ID in '/tmp/sleep_pid', and ensure the PID file is removed after the process completes.
\end{CodeBox}
\medskip
\textbf{Retrieved reflection memory.}
\begin{CodeBox}
- #1: [MEMORY TYPE] FAILURE_REFLECTION
[TASK]
Run a sleep process in the background for 3 seconds, log its start and end times (in epoch format) to '/var/log/sleep.log', store the process ID in '/tmp/sleep_pid', and ensure the PID file is removed after the process completes.

[REFLECTION]
- ROOT CAUSE: The PID file was not created because the command to capture the PID may not have executed correctly, leading to multiple executions of the sleep process without proper PID storage.
- PATTERN TO AVOID: Avoid executing commands in a loop or without proper checks, which can lead to unintended repeated executions and failures to capture necessary outputs.
- CORRECT APPROACH: Implement a structured approach with error handling and verification after each critical command to ensure that each step completes successfully before proceeding to the next.
\end{CodeBox}
\par\medskip
\textbf{Explanation.} The failure comes from missing control/verification around process management (PID capture and cleanup), which leads to repeated execution and missing artifacts. The memory transfers as a general recipe: capture PID once, verify artifacts, and sequence commands with error handling.
\par\medskip
\textbf{Target task} (retrieval rank 1; run outcome: correct).
\begin{CodeBox}
Run a sleep process in the background for 3 seconds, log its start and end times (in epoch format) to '/var/log/sleep.log', store the process ID in '/tmp/sleep_pid', and ensure the PID file is removed after the process completes.
\end{CodeBox}
\end{CaseBox}
\medskip

\begin{CaseBox}{Case Study 5 (CS5)}
\textbf{Utility.} selected=94, success=88 (93.6\%).
\par\medskip
\textbf{Origin task.}
\begin{CodeBox}
Create a secure configuration file '/etc/appconfig/settings.conf' using vi, set ownership to root:configgroup, permissions to 660, and add 'appuser' to 'configgroup'. Ensure the file contains '# Configuration file' and 'key=value' lines.
\end{CodeBox}
\medskip
\textbf{Retrieved reflection memory.}
\begin{CodeBox}
- #2: [MEMORY TYPE] FAILURE_REFLECTION
[TASK]
Create a secure configuration file '/etc/appconfig/settings.conf' using vi, set ownership to root:configgroup, permissions to 660, and add 'appuser' to 'configgroup'. Ensure the file contains '# Configuration file' and 'key=value' lines.

[REFLECTION]
- ROOT CAUSE: The initial assumption that the group `configgroup` and the user `appuser` already existed led to errors when attempting to set ownership and modify group membership.
- PATTERN TO AVOID: Avoid making assumptions about the existence of users or groups in the system without verifying first.
- CORRECT APPROACH: Always check for the existence of necessary users and groups before performing operations that depend on them, and create them if they do not exist.
\end{CodeBox}
\par\medskip
\textbf{Explanation.} This reflection encodes a high-frequency near-miss: assuming required users/groups exist. The transferable heuristic is to preflight environment state (e.g., id/getent) and create prerequisites before applying ownership/permission changes.
\par\medskip
\textbf{Target task} (retrieval rank 2; run outcome: correct).
\begin{CodeBox}
Create a configuration file '/etc/appconfig/settings.conf' using 'vi' containing the lines 'SERVER_IP=192.168.1.100' and 'DEBUG_MODE=false', set its group ownership to 'appgroup', and permissions to 640.
\end{CodeBox}
\end{CaseBox}
\medskip

\begin{CaseBox}{Case Study 6 (CS6)}
\textbf{Utility.} selected=59, success=55 (93.2\%).
\par\medskip
\textbf{Origin task.}
\begin{CodeBox}
Create a user 'testuser', add them to group 'testgroup', configure '/data' with group ownership 'testgroup' and permissions 770, create '/data/notes.txt' using 'vi' with content 'Hello from vi', and create an executable script '/check.sh' using 'vi' to verify group ownership of '/data'.
\end{CodeBox}
\medskip
\textbf{Retrieved reflection memory.}
\begin{CodeBox}
- #5: [MEMORY TYPE] FAILURE_REFLECTION
[TASK]
Create a user 'testuser', add them to group 'testgroup', configure '/data' with group ownership 'testgroup' and permissions 770, create '/data/notes.txt' using 'vi' with content 'Hello from vi', and create an executable script '/check.sh' using 'vi' to verify group ownership of '/data'.

[REFLECTION]
- ROOT CAUSE: The output being empty does not indicate a failure, but rather that the commands executed successfully without any output. 
- PATTERN TO AVOID: Assuming that an empty output signifies an error can lead to misunderstandings about command execution results. 
- CORRECT APPROACH: Always verify the success of commands by checking their exit status or using logging to confirm that actions were completed as intended.
\end{CodeBox}
\par\medskip
\textbf{Explanation.} The lesson is that empty output usually signals success, not failure; therefore success should be validated via exit status and direct state checks. This reduces false negatives across shell automation tasks.
\par\medskip
\textbf{Target task} (retrieval rank 5; run outcome: correct).
\begin{CodeBox}
Create a file '/shared/config/settings.conf' containing 'USER=testuser' and 'GROUP=testgroup', then programmatically create a user and group using the values from the file. Finally, ensure '/shared/data' is owned by the group with 770 permissions.
\end{CodeBox}
\end{CaseBox}
\medskip

\section{Prompts}
\label{app:prompts}
We provide the exact prompt strings and message templates used by our \textsc{MemRL} implementation 
across all benchmarks. To minimize ambiguity, we separate prompts used to summarize experiences into memories from
prompts used at task time for generation/inference.

\tcbset{boxsep=1.0mm}
\newtcolorbox{PromptBox}[1]{%
  enhanced,
  breakable,
  colback=white,
  colframe=black!55,
  colbacktitle=black!6,
  coltitle=black,
  fonttitle=\bfseries,
  title={#1},
  boxrule=0.5pt,
  arc=2pt,
  left=6pt,right=6pt,top=6pt,bottom=6pt,
}
\newtcblisting{PromptCode}{%
  listing engine=listings,
  listing only,
  breakable,
  colback=black!1,
  colframe=black!35,
  boxrule=0.4pt,
  arc=2pt,
  left=6pt,right=6pt,top=4pt,bottom=4pt,
  listing options={
    basicstyle=\ttfamily\small,
    columns=fullflexible,
    keepspaces=true,
    breaklines=true,
    breakatwhitespace=false,
    showstringspaces=false
  },
}

\subsection{Experience Summarization Prompts}
\label{app:memrl-prompts:experience-summarization}

\begin{PromptBox}{HLE: Experience Summarization Prompts}
\textbf{Trajectory serialization (stored as the episode trajectory).}
\begin{PromptCode}
QUESTION
{question}

SOLUTION
{model_output_stripped}
\end{PromptCode}
\medskip
\textbf{High-level script generation prompt.}
\begin{PromptCode}
Analyze the following detailed task trajectory and create a concise,
high-level script that captures the essential steps and decision points.

The script should be:
1. Generic enough to apply to similar tasks
2. Specific enough to provide useful guidance
3. 3-5 high-level steps maximum
4. Focus on the strategy and key decisions, not detailed actions

Trajectory:
{trajectory}

High-level script:
\end{PromptCode}
\medskip
\textbf{Failure reflection prompt.}
\begin{PromptCode}
Task: {task_description}

Failed trajectory:
{failed_trajectory}

This task failed. Analyze what went wrong and suggest improvements for future similar tasks.
Focus on:
1. Incorrect assumptions
2. Steps to improve
3. What to avoid next time

Provide a brief reflection:
\end{PromptCode}
\medskip
\textbf{Stored memory content templates.}
\begin{PromptCode}
# Successful memory
Task: {task_description}

SCRIPT:
{script}

TRAJECTORY:
{trajectory}

# Failure memory
TASK REFLECTION:
Task: {task_description}

What went wrong:
{reflection}

Failed approach:
{failed_trajectory}
\end{PromptCode}
\end{PromptBox}
\medskip

\begin{PromptBox}{ALFWorld: Experience Summarization Prompts}
\textbf{Trajectory serialization (stored as the episode trajectory).}
\begin{PromptCode}
The full ALFWorld dialogue history is stored as the trajectory:
  List[{"role": ..., "content": ...}, ...]
When interpolated into the summarization prompts, it is treated as a string representation.
\end{PromptCode}
\medskip
\textbf{High-level script generation prompt.}
\begin{PromptCode}
Analyze the following detailed task trajectory and create a concise,
high-level script that captures the essential steps and decision points.

The script should be:
1. Generic enough to apply to similar tasks
2. Specific enough to provide useful guidance
3. 3-5 high-level steps maximum
4. Focus on the strategy and key decisions, not detailed actions

Trajectory:
{trajectory}

High-level script:
\end{PromptCode}
\medskip
\textbf{Failure reflection prompt.}
\begin{PromptCode}
Task: {task_description}

Failed trajectory:
{failed_trajectory}

This task failed. Analyze what went wrong and suggest improvements for future similar tasks.
Focus on:
1. Incorrect assumptions
2. Steps to improve
3. What to avoid next time

Provide a brief reflection:
\end{PromptCode}
\medskip
\textbf{Stored memory content templates.}
\begin{PromptCode}
# Successful memory
Task: {task_description}

SCRIPT:
{script}

TRAJECTORY:
{trajectory}

# Failure memory
TASK REFLECTION:
Task: {task_description}

What went wrong:
{reflection}

Failed approach:
{failed_trajectory}
\end{PromptCode}
\end{PromptBox}
\medskip

\begin{PromptBox}{BCB (BigCodeBench): Experience Summarization Prompts}
\textbf{Trajectory serialization (stored as the episode trajectory).}
\begin{PromptCode}
[BCB] epoch={epoch} phase={phase} task_id={task_id}
[PROMPT]
{bcb_task_prompt}
[GENERATED CODE]
```python
{model_code}
```
[EVAL]
{"status": "...", "error": "..."}
\end{PromptCode}
\medskip
\textbf{High-level script generation prompt.}
\begin{PromptCode}
Analyze the following detailed task trajectory and create a concise,
high-level script that captures the essential steps and decision points.

The script should be:
1. Generic enough to apply to similar tasks
2. Specific enough to provide useful guidance
3. 3-5 high-level steps maximum
4. Focus on the strategy and key decisions, not detailed actions

Trajectory:
{trajectory}

High-level script:
\end{PromptCode}
\medskip
\textbf{Failure reflection prompt.}
\begin{PromptCode}
Task: {task_description}

Failed trajectory:
{failed_trajectory}

This task failed. Analyze what went wrong and suggest improvements for future similar tasks.
Focus on:
1. Incorrect assumptions
2. Steps to improve
3. What to avoid next time

Provide a brief reflection:
\end{PromptCode}
\medskip
\textbf{Stored memory content templates.}
\begin{PromptCode}
# Successful memory
Task: {task_description}

SCRIPT:
{script}

TRAJECTORY:
{trajectory}

# Failure memory
TASK REFLECTION:
Task: {task_description}

What went wrong:
{reflection}

Failed approach:
{failed_trajectory}
\end{PromptCode}
\end{PromptBox}
\medskip

\begin{PromptBox}{LLB (LifelongAgentBench): Experience Summarization Prompts}
\textbf{Trajectory serialization (stored as the episode trajectory).}
\begin{PromptCode}
{role_1}: {content_1}
{role_2}: {content_2}
...
\end{PromptCode}
\medskip
\textbf{High-level script generation prompt.}
\begin{PromptCode}
Analyze the following detailed task trajectory and create a concise,
high-level script that captures the essential steps and decision points.

The script should be:
1. Generic enough to apply to similar tasks
2. Specific enough to provide useful guidance
3. 3-5 high-level steps maximum
4. Focus on the strategy and key decisions, not detailed actions

Trajectory:
{trajectory}

High-level script:
\end{PromptCode}
\medskip
\textbf{Failure reflection prompt.}
\begin{PromptCode}
Task: {task_description}

Failed trajectory:
{failed_trajectory}

This task failed. Analyze what went wrong and suggest improvements for future similar tasks.
Focus on:
1. Incorrect assumptions
2. Steps to improve
3. What to avoid next time

Provide a brief reflection:
\end{PromptCode}
\medskip
\textbf{Stored memory content templates.}
\begin{PromptCode}
# Successful memory
Task: {task_description}

SCRIPT:
{script}

TRAJECTORY:
{trajectory}

# Failure memory
TASK REFLECTION:
Task: {task_description}

What went wrong:
{reflection}

Failed approach:
{failed_trajectory}
\end{PromptCode}
\end{PromptBox}
\medskip

\subsection{Generation and Inference Prompts}
\label{app:memrl-prompts:generation-inference}

\begin{PromptBox}{HLE: Generation and Inference Prompts}
\textbf{System prompt (exact-match).}
\begin{PromptCode}
Your response should be in the following format:
Explanation: {your explanation for your final answer}
Exact Answer: {your succinct, final answer}
Confidence: {your confidence score between 0% and 100% for your answer}
\end{PromptCode}
\medskip
\textbf{System prompt (multiple-choice).}
\begin{PromptCode}
Your response should be in the following format:
Explanation: {your explanation for your answer choice}
Answer: {your chosen answer}
Confidence: {your confidence score between 0% and 100% for your answer}
\end{PromptCode}
\medskip
\textbf{Retrieved memory injection (system message).}
\begin{PromptCode}
=== Successful Memories ===
{memory_full_content_1}

{memory_full_content_2}

=== Failed Memories (for caution) ===
{memory_full_content_3}
...
\end{PromptCode}
\medskip
\textbf{Optional repeat-attempt reflection note (system message).}
\begin{PromptCode}
You attempted this question before.
Result: {CORRECT|INCORRECT}
Question: {question}
Previous attempt (solution only):
{solution_only}
Reflect on mistakes or gaps, then solve the problem again with a better solution.
\end{PromptCode}
\medskip
\textbf{User message text block (when images exist).}
\begin{PromptCode}
Now solve the following question:

[Image IDs: {question_image_ids}]
{question}

Attached images:
1. [{img_id_1}] ({source_1})
2. [{img_id_2}] ({source_2})
...
\end{PromptCode}
\medskip
\textbf{Message ordering.}
\begin{PromptCode}
1) system: exact-match OR multiple-choice format prompt
2) system: optional reflection note (if enabled)
3) system: optional retrieved memory context
4) user: question content (text + optional images)
\end{PromptCode}
\end{PromptBox}
\medskip

\begin{PromptBox}{ALFWorld: Generation and Inference Prompts}
\textbf{Base system prompt (ReAct format + action space).}
\begin{PromptCode}
Interact with a household to solve a task. Imagine you are an intelligent agent in a household environment and your target is to perform actions to complete the task goal. At the beginning of your interactions, you will be given the detailed description of the current environment and your goal to accomplish.
For each of your turn, you will be given the observation of the last turn. You should first think about the current condition and plan for your future actions, and then output your action in this turn. Your output must strictly follow this format:"Thought: your thoughts.\nAction: your next action".

The available actions are:
1. go to {recep}
2. take {obj} from {recep}
3. move {obj} to {recep}
4. open {recep}
5. close {recep}
6. use {obj}
7. clean {obj} with {recep}
8. heat {obj} with {recep}
9. cool {obj} with {recep}
where {obj} and {recep} correspond to objects and receptacles.
After your each turn, the environment will give you immediate feedback based on which you plan your next few steps. if the envrionment output "Nothing happened", that means the previous action is invalid and you should try more options.

Your response should use the following format:

Thought: <your thoughts>
Action: <your next action>
\end{PromptCode}
\medskip
\textbf{Retrieved memory injection (system message).}
\begin{PromptCode}
In addition to the example, you have the following memories from your own past experiences. Use them to help you if they are relevant:

--- SUCCESSFUL MEMORIES (Examples to follow) ---
{formatted_success_memories_joined}

--- FAILED MEMORIES (Examples to avoid or learn from) ---
{formatted_failed_memories_joined}
\end{PromptCode}
\medskip
\textbf{Current task prompt (user message).}
\begin{PromptCode}
Now, it's your turn to solve a new task.
{task_description}
\end{PromptCode}
\medskip
\textbf{Per-step observation prompt (user message).}
\begin{PromptCode}
Observation: {observation}
\end{PromptCode}
\medskip
\textbf{Message ordering (high level).}
\begin{PromptCode}
1) system: base ALFWorld system prompt
2) user/assistant: selected few-shot example dialogue (sequence of messages)
3) system: optional retrieved memory context
4) user: new task prompt
5) loop: append user Observation: ..., model replies with Thought/Action
\end{PromptCode}
\end{PromptBox}
\medskip

\begin{PromptBox}{BCB (BigCodeBench): Generation and Inference Prompts}
\textbf{Retrieved memory injection (system message).}
\begin{PromptCode}
[Retrieved Memory Context]

### Memory 1 (id={mem_id_1}, sim={similarity_1})
{memory_content_1}

### Memory 2 (id={mem_id_2}, sim={similarity_2})
{memory_content_2}
...
\end{PromptCode}
\medskip
\textbf{Dataset-provided task prompt selection (user message).}
\begin{PromptCode}
if split == "instruct":
    return task["instruct_prompt"]
if split == "complete":
    return task["complete_prompt"]
\end{PromptCode}
\medskip
\textbf{Message ordering.}
\begin{PromptCode}
1) system: optional [Retrieved Memory Context] ...
2) user: {bcb_task_prompt}
\end{PromptCode}
\end{PromptBox}
\medskip

\begin{PromptBox}{LLB (LifelongAgentBench): Generation and Inference Prompts}
\textbf{Base system prompt.}
\begin{PromptCode}
You are an execution-focused AI agent solving database and operating-system tasks.

You may receive a [Retrieved Memory Context] block with past experiences from similar problems.
These are **references for learning**, not guaranteed solutions:
- [MEMORY TYPE] SUCCESS_PROCEDURE: A successful approach from a similar task learn the pattern.
- [MEMORY TYPE] FAILURE_REFLECTION: A failed attempt with lessons avoid similar mistakes.

Use the memories as inspiration, but always analyze your current task independently and
adapt your approach based on its specific requirements.
\end{PromptCode}
\medskip
\textbf{Strict output constraint (DB tasks).}
\begin{PromptCode}
STRICT OUTPUT FORMAT (LLB:DB, do not violate):
1) After your reasoning, include exactly ONE action line:
   - Action: Operation
   - Action: Answer
2) If Action: Operation, put exactly ONE SQL statement in the FIRST fenced code block using ```sql, on a single line. Do not add any extra text after that block.
3) If Action: Answer, include `Final Answer: ...` on the next line and do not add extra text after that.
\end{PromptCode}
\medskip
\textbf{Strict output constraint (OS tasks).}
\begin{PromptCode}
STRICT OUTPUT FORMAT (LLB:OS, do not violate):
1) After your reasoning, include exactly ONE action line:
   - Act: bash
   - Act: finish
2) If Act: bash, the next lines MUST be a ```bash fenced code block with your Bash commands. Do not include any other code blocks.
3) If Act: finish, it must be the last line (no code blocks, no extra text).
4) Do NOT use `Action:` in OS tasks (use `Act:` only).
\end{PromptCode}
\medskip
\textbf{Retrieved memory injection block.}
\begin{PromptCode}
[Retrieved Memory Context]

=== SUCCESSFUL EXPERIENCES (Learn from these) ===
[SUCCESS 1] [TYPE: {mem_type}]
{content}

=== FAILED EXPERIENCES (Avoid these mistakes) ===
[FAILURE 1] [TYPE: {mem_type}]
{content}
\end{PromptCode}
\medskip
\textbf{Prompt assembly ordering (system prompt).}
\begin{PromptCode}
1) base system prompt
2) optional [Retrieved Memory Context] ...
3) strict output format block appended at the end (task-aligned)
\end{PromptCode}
\end{PromptBox}
\end{document}